\documentclass[11pt]{article}

\usepackage[preprint]{acl}

\usepackage{times}
\usepackage{latexsym}
\usepackage[T1]{fontenc}
\usepackage[utf8]{inputenc}
\usepackage{microtype}
\usepackage{inconsolata}
\usepackage{graphicx}
\usepackage{booktabs}
\usepackage{multirow}
\usepackage{tikz}
\usepackage{todonotes}
\usepackage[greek,english]{babel}
\usepackage[utf8]{inputenc}
\usepackage{tabularx}
\usepackage{amssymb}

\newcolumntype{Y}{>{\raggedright\arraybackslash}X}

\title{GreekMMLU: A Native-Sourced Multitask Benchmark for Evaluating Language Models in Greek}

\author{
 \textbf{Yang Zhang\textsuperscript{1}$^*$},
\textbf{Mersin Konomi\textsuperscript{1}}\thanks{These authors contributed equally.},
\textbf{Christos Xypolopoulos\textsuperscript{1,3}},
\\
\textbf{Konstantinos Divriotis\textsuperscript{2}},
\textbf{Konstantinos Skianis\textsuperscript{4}},
\textbf{Giannis Nikolentzos\textsuperscript{5}},
\\
\textbf{Giorgos Stamou\textsuperscript{3}},
 \textbf{Guokan Shang\textsuperscript{2}$^\dagger$},
 \textbf{Michalis Vazirgiannis\textsuperscript{1,2}$^\dagger$}
\\
\\
 \textsuperscript{1}LIX, Ecole Polytechnique,
 \textsuperscript{2}MBZUAI,
 \textsuperscript{3}National Technical University of Athens,\\
 \textsuperscript{4}University of Ioannina,
 \textsuperscript{5}University of Peloponnese
\\
 \small{
   $^\dagger$Correspondence: \texttt{guokan.shang@mbzuai.ac.ae, mvazirg@lix.polytechnique.fr}
 }
}

\begin{document}
\maketitle

\begin{abstract}
Large Language Models (LLMs) are commonly trained on multilingual corpora that include Greek, yet reliable evaluation benchmarks for Greek—particularly those based on authentic, native-sourced content—remain limited. Existing datasets are often machine-translated from English, failing to capture Greek linguistic and cultural characteristics. We introduce GreekMMLU\footnote{\url{https://github.com/mersinkonomi/GreekMMLU}}, a native-sourced benchmark for massive multitask language understanding in Greek, comprising 21,805 multiple-choice questions across 45 subject areas, organized under a newly defined subject taxonomy and annotated with educational difficulty levels spanning primary to professional examinations. All questions are sourced or authored in Greek from academic, professional, and governmental exams. We publicly release 16,857 samples and reserve 4,948 samples for a private leaderboard\footnote{\url{https://hf.co/spaces/yangzhang33/GreekMMLU-Leaderboard}} to enable robust and contamination-resistant evaluation. Evaluations of over 80 open- and closed-source LLMs reveal substantial performance gaps between frontier and open-weight models, as well as between Greek-adapted models and general multilingual ones. Finally, we provide a systematic analysis of factors influencing performance—including model scale, adaptation, and prompting—and derive insights for improving LLM capabilities in Greek.
\end{abstract}  

\section{Introduction}

\begin{figure}[h]
  \includegraphics[width=\columnwidth]{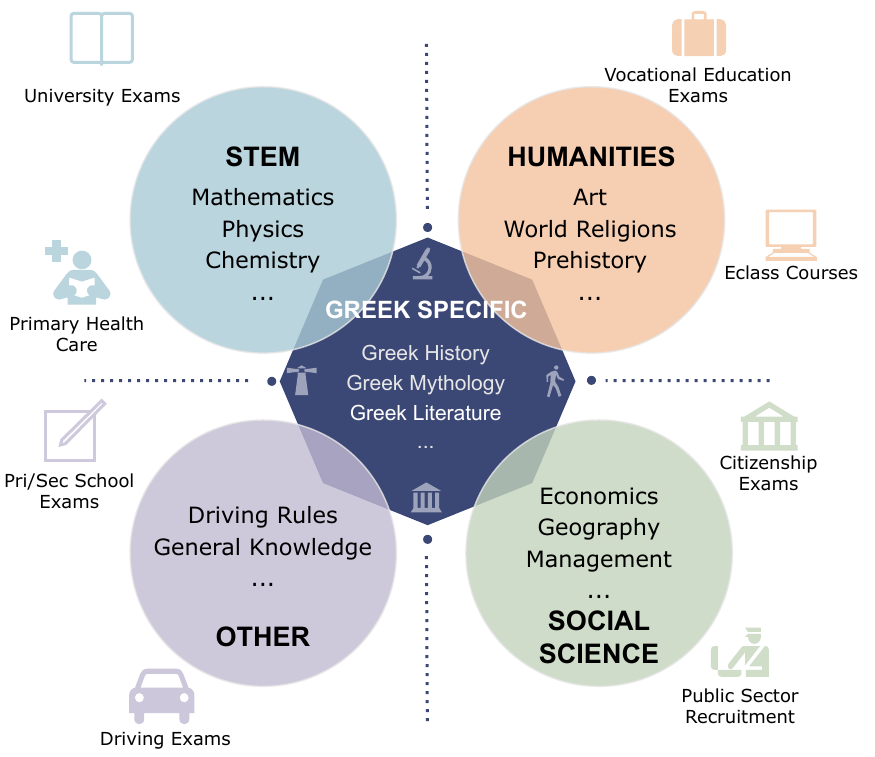}
  \caption{GreekMMLU task overview.}
  \label{fig:greekmmlu}
\end{figure}

Large language models have achieved strong performance across a wide range of natural language understanding and reasoning tasks, largely driven by large-scale training on multilingual corpora \cite{brown2020language, grattafiori2024llama, ustun2024Aya}. 
Contemporary models are designed to support dozens or even hundreds of languages, a choice driven by the need to maximize data scale and leverage cross-lingual transfer to improve general reasoning capabilities.

Furthermore, recent research has increasingly emphasized extending LLM capabilities to lower-resource languages \cite{voukoutis2024meltemi, roussis2025krikri, martins2025eurollm, shang-etal-2025-nile, shang-etal-2025-atlas}. 
In practice, while Greek is often present in the long tail of web-scale training corpora \cite{grattafiori2024llama}, it is rarely prioritized as a core language, resulting in significantly lower representation compared to major European languages.
Despite this incidental exposure, relatively little work systematically reports LLM performance on Greek, largely due to the absence of large-scale, native-language evaluation benchmarks.
Existing evaluations are commonly based on machine-translated datasets originally designed for English \cite{xuan2025mmlu, voukoutis2024meltemi}, which fail to capture the linguistic and cultural characteristics of authentic Greek language use.

A widely adopted benchmark for assessing broad knowledge and reasoning is the Massive Multitask Language Understanding (MMLU) benchmark \cite{hendrycks2020measuring}, which evaluates models across diverse subjects including STEM, social science and humanities. While MMLU has become a cornerstone of LLM evaluation, it is inherently grounded in English and the educational and cultural context of the United States. Extending MMLU to other languages through translation introduces well-known limitations, including translationese, semantic drift, altered difficulty calibration, and the preservation of source-language cultural priors \cite{singh2025global, artetxe-etal-2020-cross, li2024cmmlu}. As a result, translated benchmarks conflate native language understanding with cross-lingual transfer and often provide a misleading assessment of model capabilities.

\begin{figure*}[h]
\centering
  \includegraphics[width=\linewidth]{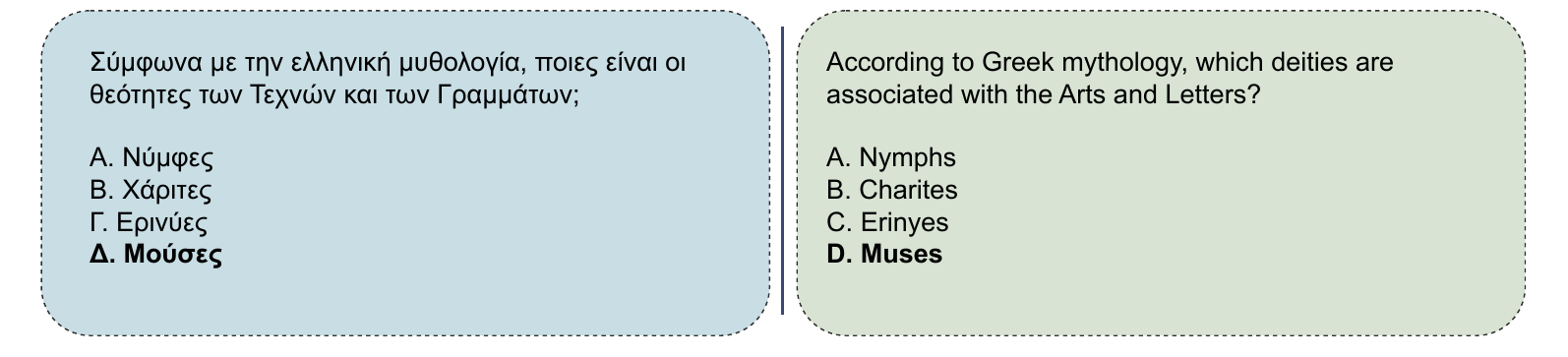}
  \caption{Example of a mythology question from GreekMMLU. \textbf{Left} shows the content structure derived from native sources and \textbf{right} is the English translation. The bold options represent the correct answer keys.}
  \label{fig:greekmmlu_history_example}
\end{figure*}

These limitations are particularly salient for Greek. Modern Greek exhibits rich morphology, flexible word order, and complex negation, all of which are difficult to evaluate through translated data. Furthermore, many knowledge-intensive domains—such as history, law, civics, and professional certification—are tightly coupled to national curricula, legal frameworks, and cultural context. Evaluating such domains using translated English benchmarks obscures important gaps in localized knowledge and overestimates real-world utility.

To address these issues, we introduce GreekMMLU, the first large-scale, native-sourced benchmark for evaluating massive multitask language understanding in Greek. GreekMMLU consists of 21{,}805 Multiple-Choice Questions (MCQ) across 45 manually defined subject areas spanning STEM, Humanities, Social Sciences, and Other domains, drawn from authentic academic, professional, and governmental examinations. All materials were collected exclusively from sources explicitly released under open-access licenses or educational reuse terms, ensuring ethical data use and legal compliance. The benchmark covers a wide range of educational and difficulty levels, including primary school, secondary school, university, and professional level, and includes eight Greek Specific subject areas requiring localized cultural knowledge. All questions are originally sourced or authored in Greek, preserving linguistic nuance and culturally grounded content.

Our contributions are summarized as follows:

\noindent$\bullet$ We introduce \textbf{GreekMMLU}, the first large-scale, fully native-sourced benchmark. It comprises 21{,}805 multiple-choice questions, organized under a \textbf{carefully defined subject taxonomy} with 45 subjects and \textbf{systematically annotated with educational difficulty levels} ranging from primary education to professional examinations.

\smallskip

\noindent$\bullet$ We conduct a large-scale evaluation of \textbf{80+ open- and closed-source LLMs} on GreekMMLU, revealing clear and consistent trends: closed-source frontier models substantially outperform open-weight alternatives, model scale strongly correlates with performance and general-purpose multilingual models exhibit pronounced weaknesses on Greek-specific and culturally grounded subject areas.

\smallskip

\noindent$\bullet$ We provide an in-depth analysis of factors influencing performance on native Greek language understanding, examining the effects of model scale, instruction tuning, prompting strategies, subject domains, and educational levels. Our findings show that Greek-adapted training leads to significant and systematic gains, particularly on tasks requiring culturally grounded knowledge.

\noindent$\bullet$ We release and maintain an official leaderboard for GreekMMLU, with separate public and private subsets to support fair, contamination-resistant evaluation.


\section{Related Work}

\paragraph{Language Models in Greek}
Greek remains underrepresented in large-scale language modeling, as most multilingual LLMs include it only implicitly and without explicit tokenizer design or data balancing, resulting in weaker performance compared to high-resource languages \cite{chowdhery2023palm, touvron2023llama}. While models such as Qwen \cite{yang2025qwen3} and Gemma \cite{team2025gemma} can perform Greek tasks, they do not explicitly report Greek training data and performance. Earlier multilingual models like BLOOMZ and mT0 \cite{muennighoff-etal-2023-crosslingual} explicitly included Greek, but only at a very limited scale (around 0.03\%). More recent efforts have addressed this gap through targeted pretraining. EuroLLM \cite{martins2025eurollm} increased coverage of European languages, including Greek. \citet{voukoutis2024meltemi} introduced \textit{Meltemi}, the first openly released Greek-centric LLM, trained with large-scale Greek corpora and a Greek-aware tokenizer, achieving substantial gains on Greek benchmarks. Building on this work, \citet{roussis2025krikri} proposed \textit{LLaMA-Krikri}, further expanding Greek coverage through increased Greek data. Beyond general-purpose models, domain-specific efforts such as \textit{Plutus-8B} demonstrate the benefits of Greek-adapted training in specialized settings \cite{peng2025plutus}. Overall, these works highlight the importance of explicit Greek-focused training for robust Greek language understanding.

\paragraph{LLM Evaluation and Multilingual Benchmarks}
General knowledge and reasoning in LLMs are commonly evaluated using multitask benchmarks such as MMLU \cite{hendrycks2020measuring}, along with datasets like ARC \cite{clark2018think} and HellaSwag \cite{zellers2019hellaswag}. While effective for tracking architectural and scaling progress, benchmarks like MMLU are predominantly English-centric, limiting their ability to assess culturally grounded language understanding. Multilingual extensions often rely on machine translation, including multilingual variants of MMLU \cite{xuan2025mmlu}, but prior work has shown that translated benchmarks suffer from translationese, semantic drift, altered difficulty calibration, and inherited cultural priors, leading to validity concerns \cite{singh2025global, artetxe-etal-2020-cross, li2024cmmlu}. To address these issues, native-sourced benchmarks such as CMMLU for Chinese \cite{li2024cmmlu}, ArabicMMLU for Arabic \cite{koto2024arabicmmlu}, and TurkishMMLU for Turkish \cite{yuksel2024turkishmmlu} have been proposed, highlighting the need for locally grounded multilingual evaluation.

\paragraph{Greek LLM Evaluation}
Evaluation of Greek language capabilities has traditionally relied on multilingual benchmarks such as XNLI \cite{conneau-etal-2018-xnli}, XTREME \cite{hu2020xtreme}, XQuAD \cite{artetxe-etal-2020-cross}, MASSIVE \cite{fitzgerald-etal-2023-massive}, and FLORES-101 \cite{goyal-etal-2022-flores}, where Greek is included as a target language to assess cross-lingual generalization. 
More recent Greek-centric efforts, such as Meltemi \cite{voukoutis2024meltemi} and KriKri \cite{roussis2025krikri}, largely resorted to machine-translated versions of English benchmarks like HellaSwag and MMLU. 
Several Greek-specific benchmarks enable a more targeted evaluation. GreekSUM \cite{evdaimon-etal-2024-greekbart} introduced the first large-scale abstractive summarization dataset for Greek news. Earlier resources such as elNER \cite{bartziokas2020datasets}, support named entity recognition in Greek, and recent domain-specific benchmarks like Plutus-ben \cite{peng2025plutus} and GreekBarBench \cite{chlapanis2025greekbarbench} extend evaluation to financial and legal reasoning tasks. However, these efforts are typically task-specific and limited in scale, reflecting a need for dedicated benchmarks that better capture the linguistic and cultural specific challenges in Greek.

\section{The GreekMMLU Dataset}

\subsection{Overview}
\label{sec:overview}

GreekMMLU is a \textbf{native-sourced benchmark} for evaluating massive multitask language understanding in Greek, composed exclusively of original Greek questions drawn from real-world educational and professional assessments. All questions follow an MCQ format as illustrated in Figure \ref{fig:greekmmlu_history_example}, with a variable number of answer options (2--4) and exactly one correct answer,  reflecting the diversity of Greek national examinations. Each question is annotated with a difficulty level corresponding to its educational context.

A major effort in creating GreekMMLU involved the systematic structuring of heterogeneous raw exam material. We designed a \textbf{custom subject taxonomy} and \textbf{carefully assigned each task to an educational difficulty level}, enabling consistent analysis across domains and degrees of specialization. The resulting benchmark spans 45 subject areas, organized into four high-level categories---\textbf{STEM, Humanities, Social Sciences, and Other} as shown in Figure \ref{tab:subject_groups}, covering educational levels from \textbf{primary school to university and professional examinations}.

Beyond general academic content, GreekMMLU includes a dedicated subset of \textbf{Greek-specific tasks} that require explicitly Greek linguistic and cultural knowledge, such as Greek History, Greek Literature, Greek Mythology, Greek Traditions, and the Modern Greek Language. These tasks are tightly coupled to the local cultural context and cannot be reliably assessed through translated benchmarks, highlighting the importance of native-sourced evaluation.

The dataset is divided into a \textbf{public, open-source subset} released for research use and a \textbf{private subset} reserved for a leaderboard to support more robust and contamination-resistant evaluation. Representative examples of the dataset are provided in Appendix~\ref{example}.




\begin{table}[t]
\centering
\footnotesize
\renewcommand{\arraystretch}{1.15}
\setlength{\tabcolsep}{2pt}

\begin{tabular}{l >{\raggedright\arraybackslash
                   \hyphenpenalty=10000
                   \exhyphenpenalty=10000}p{0.70\columnwidth}}
\toprule
\textbf{Group} & \textbf{Subjects} \\
\midrule

Humanities &
Art (S, U, Pr), Greek History (P, S, Pr), Greek Literature (S, U), Greek Mythology (P, S, U), Law (S, Pr), Prehistory (P), World History (P, S), World Religions (S, U) \\
\midrule

STEM &
Agriculture (U, Pr), Biology (P, S), Chemistry (P, U), Civil Engineering (Pr), Clinical Knowledge (Pr), Computer Networks \& Security (U), Computer Science (U, Pr), Electrical Engineering (U, Pr), Mathematics (P, U), Medicine (U, Pr), Physics (P, U, Pr) \\
\midrule

Social Sciences &
Economics (U, Pr), Education (U, Pr), Geography (P, S), Government and Politics (P, S), Greek Traditions (S, Pr), Management (U, Pr), Modern Greek Language (P, S),Accounting (Pr) \\
\midrule

Other &
 Driving Rules (NA), General Knowledge (S, Pr), Maritime Safety and Rescue Operations (Pr) \\

\bottomrule
\end{tabular}

\caption{Subject areas in GreekMMLU. ``P'', ``S'', ``U'', ``Pr'', and ``NA'' indicate availability in primary school, secondary school, university, professional, and not available categories, respectively.}
\label{tab:subject_groups}
\end{table}

\subsection{Data Collection and Curation}
We conducted an exhaustive survey of publicly available Greek testing platforms to compile a corpus of questions and answers. 
Our sourcing strategy targeted authoritative bodies, identifying a wide spectrum of standardized examinations ranging from primary education to professional licensure.
We systematically crawled and ingested data from these repositories, developing custom extraction pipelines to handle heterogeneous file formats—including structured web interfaces, PDF archives, and DOCX documents. 
These formats typically reside outside the scope of standard web-crawling pipelines (e.g., Common Crawl) used for LLM pre-training, thereby minimizing the risk of data contamination.
This process yielded a raw corpus of diverse subject matter, ensuring the benchmark captures the breadth of the Greek educational and professional curriculum.

Given the predominance of PDF documents in the raw corpus, we utilized \textit{PyMuPDF4LLM}\footnote{\url{https://github.com/pymupdf/pymupdf4llm}} for structure-aware text extraction. For legacy scanned files, we applied Tesseract OCR\footnote{\url{https://github.com/tesseract-ocr/tesseract}}. To mitigate extraction artifacts—particularly in malformed Greek text and mathematical notations—we implemented an LLM-assisted correction phase utilizing Claude 3.5 Sonnet and Haiku. 
The models were prompted to preserve the original semantic intent verbatim and identify toxic content, ensuring that no generative question creation occurred during the process.

Following this automated restoration, the dataset underwent strict Unicode canonicalization and punctuation standardization (e.g., correcting the Greek question mark '$;$'). Finally, the curated dataset---after filtering and masking any personal identifying information and meaningless ids---was validated by a dedicated team of \textbf{five native Greek-speaking experts} holding graduate-level academic qualifications. This team manually reviewed the question--answer pairs to verify linguistic fidelity and filter out remaining processing artifacts, ensuring the benchmark adheres to the rigorous standards of authentic Greek examinations.

\subsection{Quality Control}
In constructing GreekMMLU, the dataset was derived predominantly from official sources, whose data quality was assumed to be reliable based on their authoritative provenance. Materials from unofficial or secondary sources constituted 26.2\% of the corpus (approximately 4,400 samples) and were therefore subjected to comprehensive human review. Each of these samples was manually inspected by our experts, resulting in the identification and correction of errors in 3.8\% of the subset. To further ensure minimal residual noise introduced during OCR and parsing stages, we conducted an additional round of expert human validation on a randomly sampled set of 8,000 examples drawn across all sources. This multi-stage verification pipeline underscores the substantial human effort invested in ensuring the overall accuracy and robustness of GreekMMLU.

For the final dataset, we performed a rigorous human evaluation. Three Greek-speaking graduate- and professor-level experts randomly selected \textbf{5\%} of the samples from each individual task and manually verified both the question stems and the corresponding ground-truth answers. This evaluation focused on identifying residual OCR errors, semantic drift, or incorrect labeling. Based on this process, we estimated the overall noise level in the dataset—the proportion of QA samples deemed low quality or incorrect—to be approximately \textbf{2\%}, with all identified issues corrected prior to release.

\begin{table}[h]
\centering
\small
\setlength{\tabcolsep}{5pt}
\begin{tabular}{lrrr}
\toprule
\multirow{2}{*}{\textbf{Group}} &
\multirow{2}{*}{\textbf{\# Questions}} &
\multicolumn{2}{c}{\textbf{\# Chars}} \\
\cmidrule(lr){3-4}
 & & \textbf{Question} & \textbf{Answer} \\
\midrule
STEM & 6787 & 94.5 & 26.5 \\
Humanities & 2751 & 93.8 & 42.5 \\
Social Sciences & 4753 & 306.4 & 21.4 \\
Other & 2341 & 86.4 & 51.9 \\
\midrule
Primary School & 4393 & 67.2 & 19.9 \\
Secondary School & 1566 & 747.2 & 19.7 \\
Professional & 8103 & 106.5 & 33.0 \\
University & 912 & 90.7 & 42.2 \\
NA & 1658 & 88.7 & 57.6 \\
\bottomrule
\end{tabular}
\caption{Average question and answer length (in characters) for each education group and subject area in GreekMMLU.}
\label{tab:subject_stats}
\end{table}

\subsection{Statistics and Analysis}
GreekMMLU consists of \textbf{21,805} multiple-choice questions across \textbf{45 subjects}, organized into \textbf{four supercategories} (STEM, Humanities, Social Sciences, and Other) and annotated with \textbf{four difficulty levels} corresponding to \textit{Primary}, \textit{Secondary}, \textit{University}, and \textit{Professional} education. We split about 23\% from each subject to create two subsets. The \textbf{public subset contains 16,857 questions} and is used for all experiments reported in this work, while the \textbf{private subset contains 4,948 questions} and is reserved for a future evaluation leaderboard. A detailed statistical distribution can be found in Table \ref{tab:subject_stats} and Appendix \ref{statistic}.

As shown in Table \ref{tab:subject_stats}, most questions are drawn from professional examinations, followed by primary, secondary, and university-level questions, with an additional NA category of 1.7K questions not aligned to a specific educational tier. Average question and answer length generally increases with educational level; secondary school items form a notable exception, with longer prompts but relatively short answers, reflecting the inclusion of citizenship and civic education tasks that combine context-rich descriptions with concise answer options. Across subject areas, the dataset is broadly balanced, with STEM, humanities, and social sciences each comprising 2.7K--6.8K questions; social science questions are longer on average due to text-based, situational prompts, while STEM questions tend to be more concise and formula-driven.

\section{Experiments and Results}


\subsection{Experimental Setup}

We integrate GreekMMLU into the lm-evaluation-harness framework \cite{eval-harness} to ensure a standardized and reproducible evaluation environment. All models are evaluated under both zero-shot and five-shot settings. We show the detailed setup in Appendix \ref{sec:implementation_details}.

\begin{figure}[h]
  \includegraphics[width=\columnwidth]{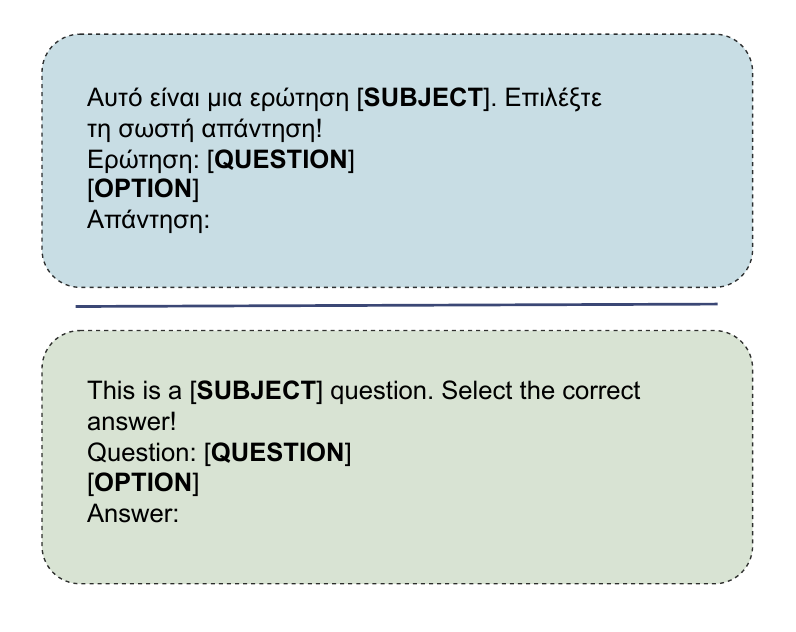}
  \caption{Prompt templates in Greek and English.}
  \label{fig:prompt_templates}
\end{figure}

\paragraph{\textbf{Evaluation Strategy.}}
Evaluation depends on access to model outputs. For open-weights models, we use a rank-based multiple-choice setup, selecting the answer with the highest token log-likelihood. For closed-source API models, we use free-form generation, prompting models to output the answer key directly and extracting the predicted Greek label with regular expressions \cite{li2024cmmlu}.

\paragraph{\textbf{Prompting Protocol.}}
Models are evaluated under both zero-shot and five-shot prompting using prompts written entirely in Greek. Each prompt includes a subject-specific instruction, the question stem, and the labeled answer options, following Greek examination conventions with answer labels A, B, $\Gamma$, and $\Delta$. In the five-shot setting, five representative examples from the task’s development set are prepended to the prompt. The prompt templates are illustrated in Figure \ref{fig:prompt_templates}.

\begin{table*}[p]
\centering
\footnotesize
\begin{tabular}{l cccc | c| c}
\toprule
\textbf{Model} & \textbf{STEM} & \textbf{Humanities} & \textbf{Social Sci.} & \textbf{Other} & \textbf{Average} & \textbf{Greek-specific} \\
\midrule
\multicolumn{7}{l}{\textbf{General-Purpose LLMs}} \\
\midrule

GPT-5.2$^\dagger$  & 86.05 & 88.27 & 90.29 & 85.96 & 87.75 & 92.92 \\
GPT-4o$^\dagger$   & 84.54 & 88.68 & 89.36 & 85.42 & 86.81 & 93.11 \\
Gemini 3 Flash$^\dagger$ & 92.82 & 92.88 & 94.16 & 91.84 & 93.16 & 95.44 \\
\addlinespace[0.5em]

Qwen2.5-7B                   & 51.92 & 56.37 & 59.34 & 53.31 & 55.16 & 61.07 \\
Qwen2.5-7B-Instruct$^\dagger$ & 59.88 & 58.33 & 61.98 & 58.89 & 60.25 & 64.02 \\
Qwen2.5-14B                  & 64.11 & 64.95 & 66.04 & 60.23 & 64.39 & 67.95 \\
Qwen2.5-14B-Instruct$^\dagger$ & 65.20 & 66.41 & 69.78 & 62.92 & 66.61 & 73.06 \\
Qwen2.5-32B                  & 72.74 & 71.29 & 75.26 & 70.68 & 73.14 & 77.40 \\
Qwen2.5-32B-Instruct$^\dagger$ & 72.08 & 71.47 & 76.61 & 69.69 & 73.22 & 80.03 \\
Qwen2.5-72B                  & 78.31 & 78.78 & 81.35 & 76.75 & 79.20 & 83.91 \\
Qwen2.5-72B-Instruct$^\dagger$ & 78.90 & 79.14 & 81.92 & 76.95 & 79.70 & 84.67 \\
Qwen3-30B                    & 70.72 & 69.06 & 74.95 & 59.63 & 70.56 & 77.49 \\
Qwen3-30B-Instruct$^\dagger$ & 79.31 & 74.81 & 79.33 & 76.56 & 78.39 & 81.80 \\

\addlinespace[0.5em]
Llama-2-7b-hf                & 36.23 & 35.97 & 33.80 & 35.54 & 35.30 & 32.84 \\
Llama-2-7b-chat-hf$^\dagger$ & 36.63 & 34.92 & 34.26 & 33.85 & 35.27 & 33.61 \\

Llama-3.1-8B                 & 51.08 & 57.42 & 54.78 & 50.62 & 53.10 & 54.78 \\
Llama-3.1-8B-Instruct$^\dagger$ & 56.58 & 62.85 & 62.56 & 57.84 & 59.56 & 64.75 \\
Llama-3.1-70B                & 72.42 & 79.05 & 78.90 & 74.46 & 75.71 & 82.46 \\

Llama-3.2-1B                 & 37.37 & 36.51 & 34.86 & 36.83 & 36.35 & 33.66 \\
Llama-3.2-1B-Instruct$^\dagger$ & 38.01 & 37.33 & 36.87 & 35.94 & 37.29 & 35.46 \\

Llama-3.2-3B                 & 41.82 & 41.67 & 43.81 & 44.75 & 42.82 & 41.97 \\
Llama-3.2-3B-Instruct$^\dagger$ & 43.77 & 43.04 & 45.62 & 45.64 & 44.52 & 46.15 \\

Llama-3.3-70B-Instruct$^\dagger$ & 77.03 & 82.20 & 82.92 & 76.80 & 79.65 & 86.94 \\

\addlinespace[0.5em]
Gemma-3-4B-pt                & 50.77 & 52.67 & 53.83 & 52.17 & 52.21 & 55.00 \\
Gemma-3-4B-it$^\dagger$      & 59.79 & 60.43 & 65.95 & 62.32 & 62.24 & 68.58 \\

Gemma-3-12B-pt               & 73.30 & 74.44 & 78.28 & 72.27 & 74.99 & 80.71 \\
Gemma-3-12B-it$^\dagger$     & 72.95 & 75.13 & 79.28 & 72.62 & 75.31 & 82.21 \\

Gemma-3-27B-pt               & 77.81 & 79.83 & 80.52 & 76.95 & 78.88 & 83.33 \\
Gemma-3-27B-it$^\dagger$     & 78.03 & 79.87 & 82.19 & 75.96 & 79.41 & 85.33 \\

\addlinespace[0.5em]
Aya-101$^\dagger$            & 52.62 & 53.75 & 62.16 & 57.05 & 56.73 & 59.86 \\
BLOOMZ-7b1$^\dagger$         & 34.27 & 32.63 & 30.03 & 32.30 & 32.40 & 28.80 \\
mT0-xxl$^\dagger$            & 52.72 & 53.13 & 61.33 & 36.92 & 56.57 & 56.91 \\
GLM-4-9b                     & 63.19 & 63.81 & 68.22 & 63.41 & 64.98 & 68.52 \\
GLM-4-9b-chat$^\dagger$      & 61.44 & 64.26 & 68.42 & 65.85 & 64.68 & 69.29 \\

\midrule
\multicolumn{7}{l}{\textbf{Greek and European LLMs}} \\
\midrule
Llama-Krikri-8B-Base         & 59.83 & 68.92 & 65.56 & 62.92 & 63.33 & 68.72 \\
Llama-Krikri-8B-Instruct$^\dagger$ & 62.62 & 70.29 & 70.57 & 64.11 & 66.47 & 74.73 \\

Meltemi-7B-v1.5              & 52.13 & 55.23 & 53.74 & 52.36 & 53.11 & 56.28 \\
Meltemi-7B-Instruct-v1.5$^\dagger$ & 57.18 & 63.99 & 64.31 & 61.03 & 60.93 & 66.42 \\

Plutus-8B-instruct$^\dagger$ & 61.65 & 69.74 & 69.73 & 64.01 & 65.71 & 73.96 \\

EuroLLM-1.7B                 & 33.21 & 34.28 & 30.54 & 32.11 & 32.33 & 29.86 \\
EuroLLM-1.7B-Instruct$^\dagger$ & 29.47 & 30.76 & 28.76 & 31.76 & 29.68 & 30.16 \\

EuroLLM-9B                   & 62.52 & 71.11 & 69.46 & 65.21 & 66.30 & 73.74 \\
EuroLLM-9B-Instruct$^\dagger$ & 64.42 & 73.16 & 72.24 & 66.85 & 68.48 & 76.86 \\

EuroLLM-22B                  & 66.94 & 75.35 & 73.49 & 68.44 & 70.43 & 77.46 \\
EuroLLM-22B-Instruct-2512$^\dagger$ & 69.78 & 75.31 & 74.66 & 70.08 & 72.18 & 78.99 \\

\midrule
\textbf{Random Baseline} & 32.33 & 28.77 & 31.86 & 32.62 & 30.42 & 31.59 \\
\bottomrule
\end{tabular}
\caption{Overall zero-shot performance of different models on the GreekMMLU benchmark. Accuracy (\%) is reported. Models marked with$^\dagger$ are instruction-tuned.}
\label{tab:main_results_combined}
\end{table*}

\paragraph{\textbf{Model Selection.}}

We conduct a comprehensive analysis of diverse model families on the GreekMMLU benchmark. Our study covers three categories. We first establish broad capabilities using \textbf{General-Purpose LLMs}, including the Qwen \cite{yang2025qwen3}, Llama \cite{grattafiori2024llama}, and Gemma-3 \cite{team2025gemma}, GLM-4 \cite{glm} families. We compare these against \textbf{Greek-Adapted LLMs}—such as Meltemi \cite{voukoutis2024meltemi}, Krikri \cite{roussis2025krikri}, and Plutus \cite{peng2025plutus}—which are selected to quantify the benefits of native-language specialization. Additionally, we evaluate \textbf{Multilingual European LLMs}, represented by the EuroLLM family \cite{martins2025eurollm} alongside multilingual models like Aya-101 \cite{ustun2024Aya}, BLOOMZ \cite{muennighoff-etal-2023-crosslingual}, and mT0 \cite{muennighoff-etal-2023-crosslingual}. Finally, we assessed frontier closed-source models like ChatGPT and Gemini, while a \textbf{Random Baseline} is included to establish a theoretical lower bound.

\subsection{Main Results}
\label{sec:main-results}

The overall zero-shot performance on part of our evaluated models is summarized in Table~\ref{tab:main_results_combined}. We show all the evaluated models with both zero-shot and five-shot settings in Appendix \ref{experiment_results}. In this paper, we only report results on the public subset, we provide analysis for the private results in Appendix \ref{corrpubpri}.

Model performance on GreekMMLU spans a wide spectrum, ranging from near-random accuracy for weak models to strong results for state-of-the-art models. Small and lightly adapted models have performance near the random baseline, while mid-sized models (3-20B) typically achieve moderate performance in the 55--70\% range. A clear separation emerges at the top end, where closed-source frontier models substantially outperform all open-weight alternatives. For example, \textit{Gemini~3~Flash} reaches an average accuracy of 93.16\%, while \textit{GPT-5.2} and \textit{GPT-4o} achieve 87.75\% and 86.81\%, respectively, consistently excelling in all subjects. In contrast, the strongest open-weight models---such as \textit{Llama-3.3-70B-Instruct} and \textit{Qwen2.5-72B-Instruct}---peak at 79.56\% and 79.70\% average accuracy, leaving a substantial gap to the best models. This persistent performance margin reflects the advantages of closed-source models, including broader exposure to Greek data during training in addition to large-scale optimization and advanced alignment.

\begin{figure}[h]
  \includegraphics[width=\columnwidth]{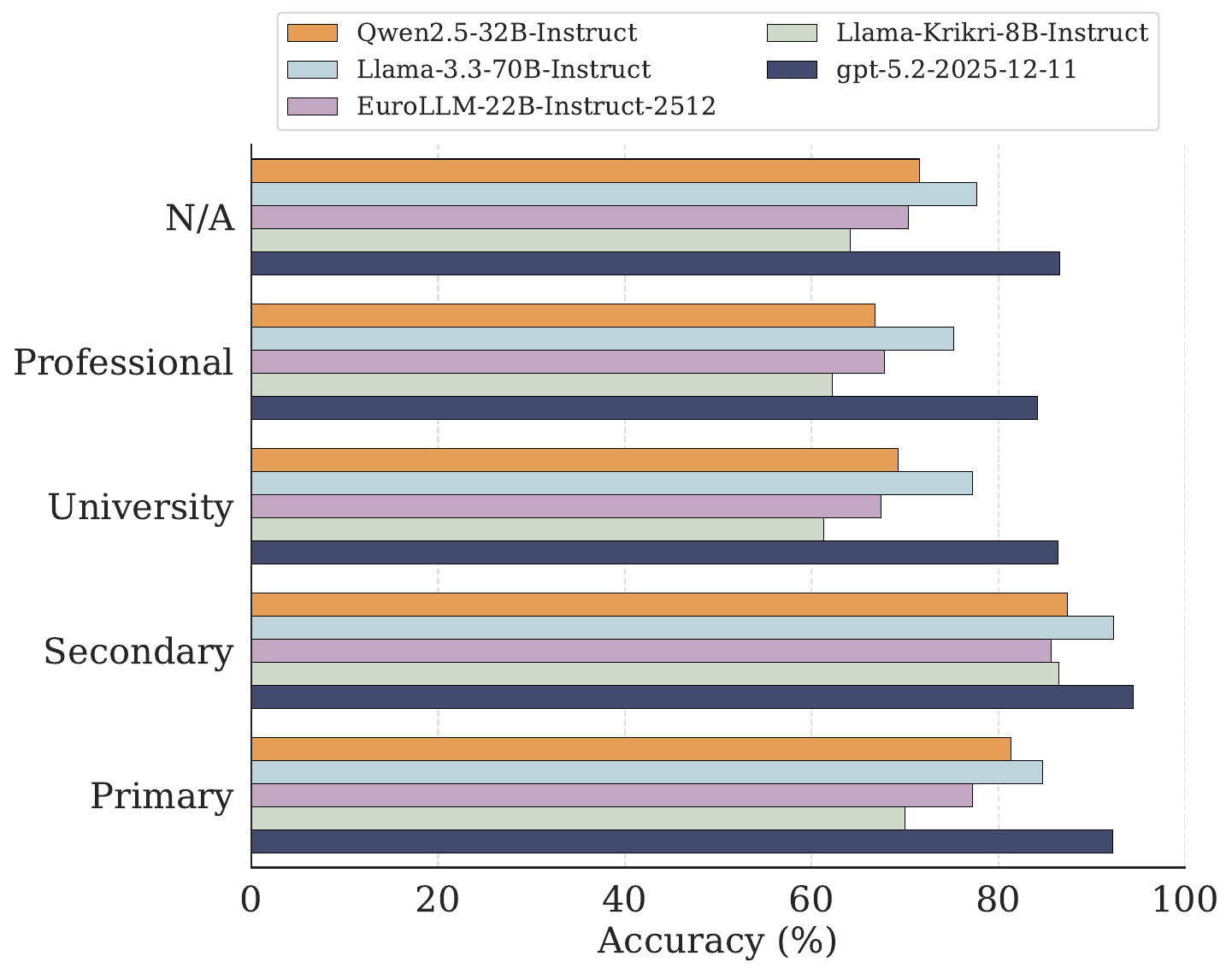}
  \caption{Models' 0-shot performance on different task levels.}
  \label{fig:level_performance}
\end{figure}

\paragraph{Instruction Tuning Effects.}
Across nearly all model families, instruction-tuned variants consistently outperform their corresponding base models at similar parameter scales. This improvement is observed across subject categories and is particularly pronounced on Greek-specific tasks. Instruction tuning appears to improve alignment with the multiple-choice evaluation format, Greek prompt structure, and answer selection conventions used in GreekMMLU.

\paragraph{Greek- and European-Centric Models.}
Models explicitly adapted to Greek or European languages consistently achieve higher accuracy than general multilingual baselines of comparable size. Greek-centric models like LLaMA-Krikri-8B demonstrate notable gains, especially on Greek-specific subjects. Similarly, European-centric models outperform globally trained multilingual models, indicating that regional linguistic focus contributes meaningfully to Greek language understanding.

\paragraph{Performance Across Educational Levels.}
Figure \ref{fig:level_performance} shows that accuracy is generally higher on primary and secondary school questions and lower on university and professional-level tasks. This pattern appears across model families and scales, including both open-weight and closed-source models. Performance on N/A-level questions typically lies between primary/secondary and university/professional levels. Overall, the reduced accuracy at higher levels reflects the increased complexity, specialized knowledge of university and professional examination questions.

\begin{figure}[h]
  \includegraphics[width=\columnwidth]{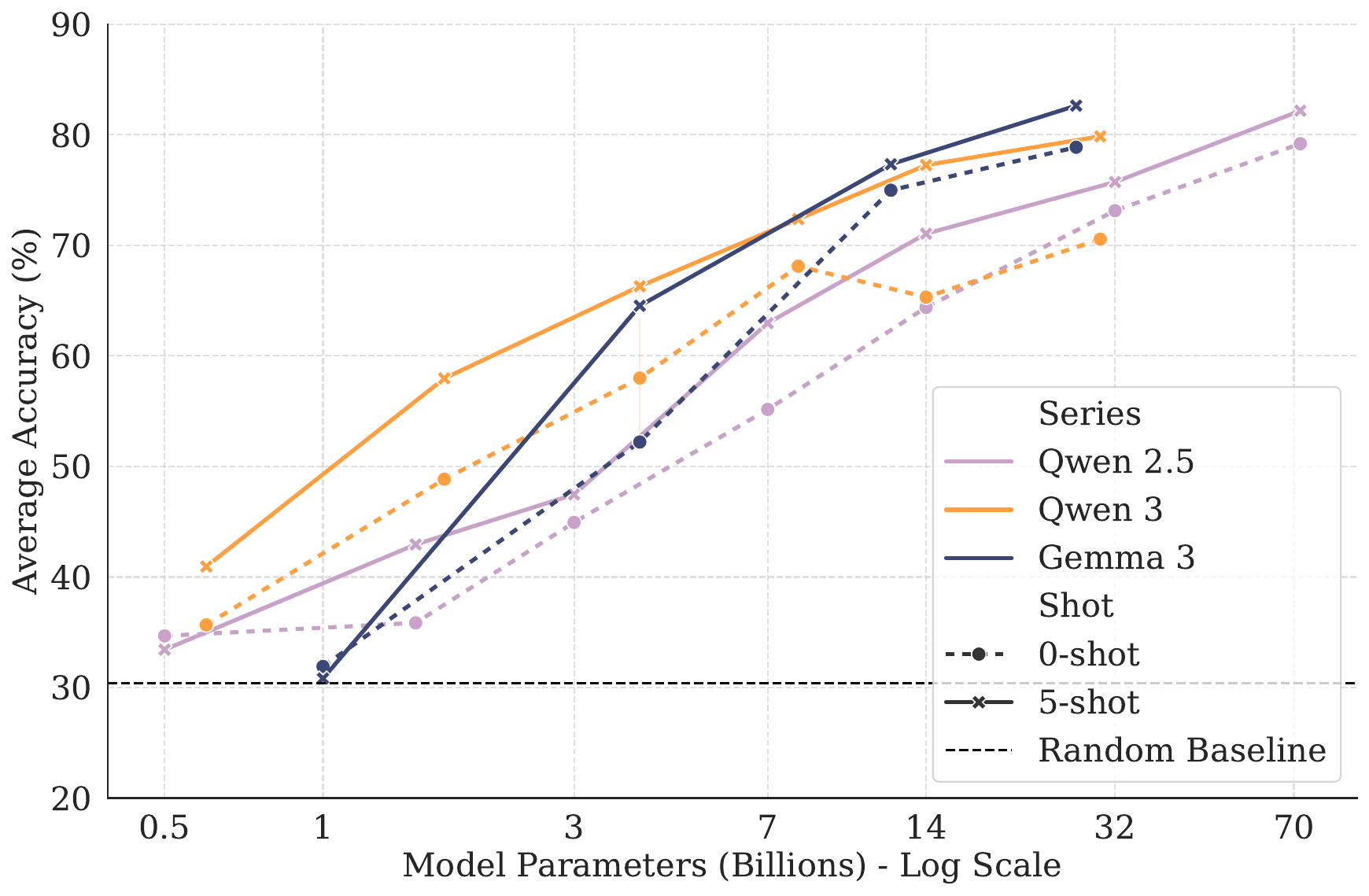}
  \caption{Scaling behavior of average accuracy with respect to model size (in billions of parameters) under zero-shot and five-shot prompting.}

  \label{fig:scaling_law}
\end{figure}

\subsection{Analysis}
\label{sec:analysis}




\paragraph{Model Scale Effects.}
Figure \ref{fig:scaling_law} shows a clear relationship between model size and performance on GreekMMLU. Very small models (below approximately 2B parameters) generally cannot solve these tasks, with accuracy close to the random baseline across all model families. As model size increases, performance improves consistently. Larger models achieve higher accuracy, reflecting better answer selection and more effective handling of longer and more complex question contexts in Greek. This trend is observed across all evaluated model series, with steady gains as parameters increase, indicating that larger models are better equipped to handle the linguistic and knowledge demands of the benchmark.

\paragraph{Subject-Level and Cultural Differences.}
Model performance varies across subject domains, with most models achieving higher accuracy in humanities and social sciences than in STEM, reflecting the greater reasoning and technical demands of STEM questions in GreekMMLU. Greek-specific subjects (e.g., history, traditions, mythology) are consistently more challenging than globally shared domains, particularly for general-purpose multilingual models. In contrast, Greek- and European-centric models of comparable size perform better on these culturally grounded tasks, indicating stronger coverage of localized knowledge. Additional analysis is provided in Appendix~\ref{sec:subject_distribution}.

\begin{figure}[h]
  \includegraphics[width=\columnwidth]{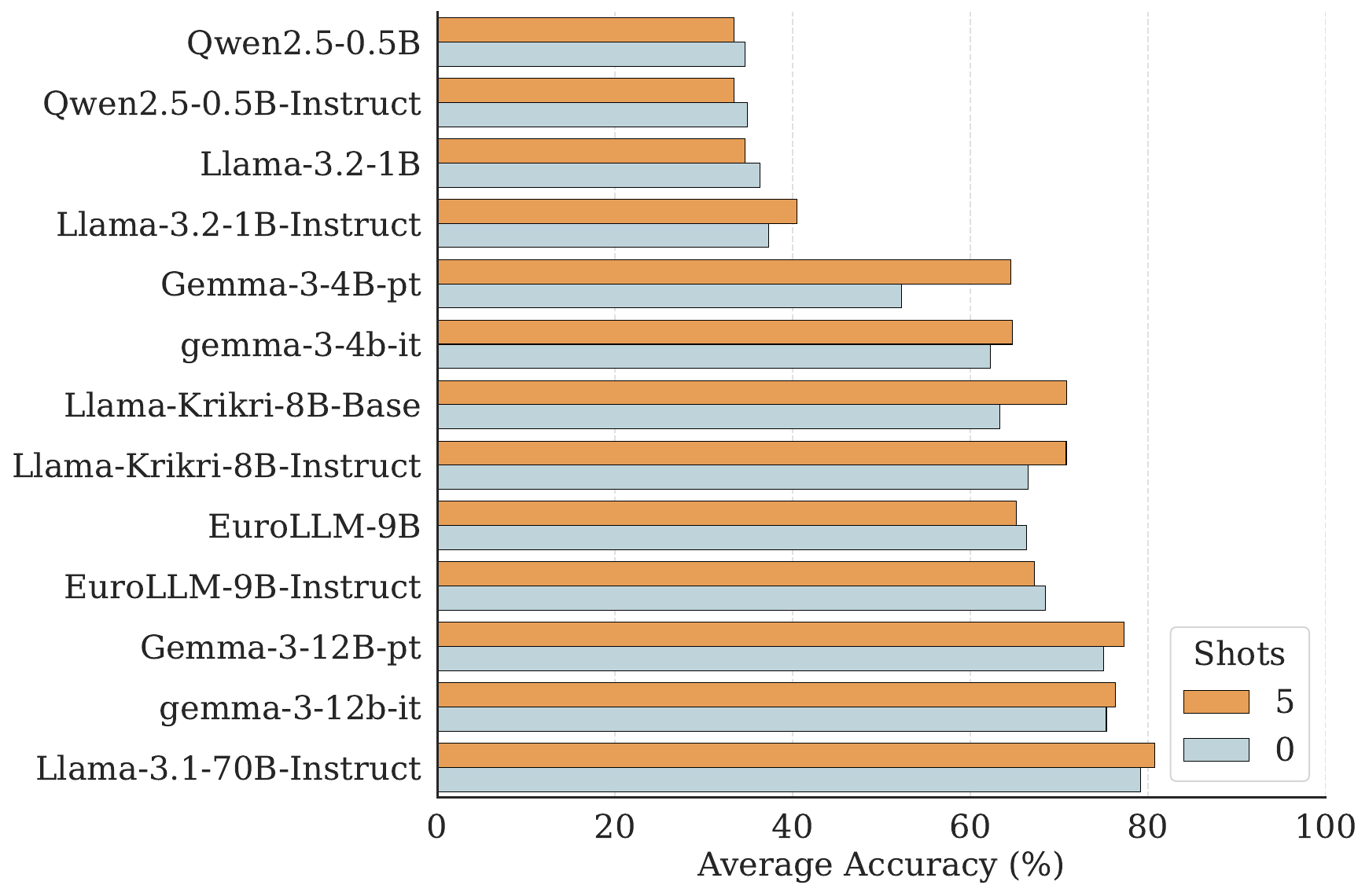}
  \caption{Comparison of average accuracy across models under zero-shot and five-shot prompting.}
  \label{fig:shot_comparison_barplot}
\end{figure}

\paragraph{Zero-Shot vs.\ Five-Shot Performance.}
Figure \ref{fig:shot_comparison_barplot} shows that five-shot prompting does not improve performance for models less than 2B, which remain close to the random baseline. In contrast, larger models consistently benefit from additional in-context examples. The improvement is most pronounced for Gemma and Llama families, where five-shot prompting yields noticeable accuracy boosts. Overall, the effectiveness of five-shot prompting is positive and can provide meaningful gains for mid- and large-scale ones.


\begin{figure}[h]
  \includegraphics[width=\columnwidth]{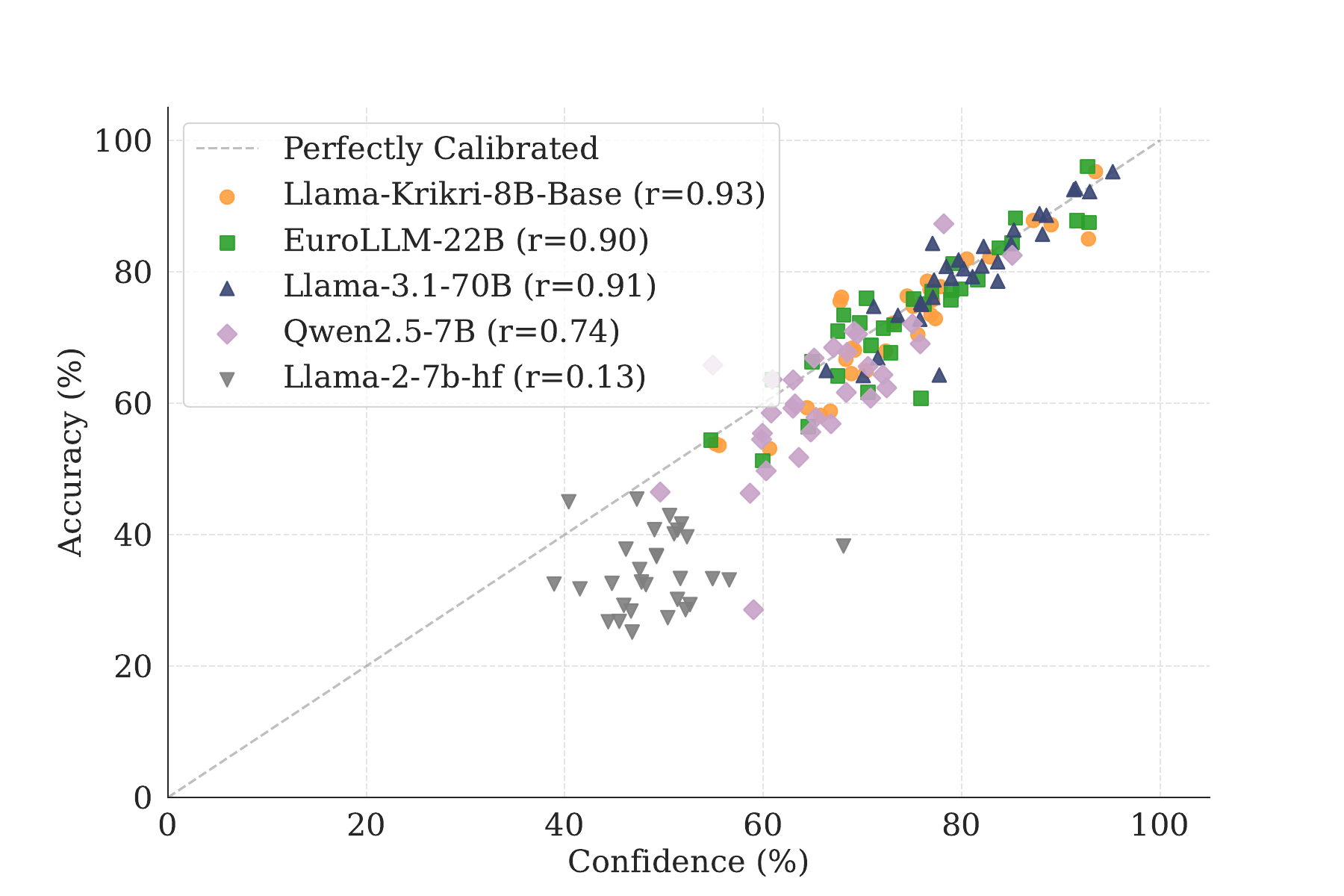}
  \caption{Calibration behavior across models on Greek-MMLU. }
  \label{fig:accuracy_vs_confidence}
\end{figure}

\paragraph{Calibration Analysis }

Our analysis of subject-level calibration on 5-shot results reveal notable differences across model families (Figure \ref{fig:accuracy_vs_confidence}). The Greek-specialized Llama-Krikri-8B shows strong alignment between confidence and accuracy (r = 0.93), while generic multilingual models such as Qwen-2.5-7B are less well calibrated (r = 0.74). Earlier-generation models (e.g., Llama-2-7B) exhibit pronounced miscalibration (r = 0.13). Although larger models (e.g., Llama-3.1-70B) reduce calibration errors, language-specific training remains the primary driver of reliable calibration. We provide extended calibration analysis in Appendix \ref{calib}.

\paragraph{Correlation Between Question Length and Model Confidence}
We further experiment with the correlation between the question length and model prediction confidence in Appendix \ref{sec:length_vs_confidence} across different model families and sizes. We found that the model confidence has little or no correlation with question length.

\section{Conclusion}
We introduced GreekMMLU, the first large-scale, native-sourced benchmark for evaluating massive multitask language understanding in Greek. GreekMMLU comprises 21,805 multiple-choice questions across 45 subjects and multiple educational levels, enabling linguistically and culturally grounded evaluation beyond machine-translated benchmarks. Our comprehensive evaluation highlights substantial performance gaps across model families and demonstrates the benefits of instruction tuning and Greek-adapted training, particularly on Greek-specific domains. We hope GreekMMLU encourages future models to place greater emphasis on native Greek capability and supports the development of more authentic, culturally grounded Greek language models.
 

\section*{Limitations}
GreekMMLU is limited to multiple-choice questions drawn from formal educational and professional settings, enabling standardized evaluation but not capturing open-ended generation, interactive reasoning, or informal language use. Although all content is natively sourced in Greek, the benchmark primarily reflects standard Modern Greek as used in official curricula, with limited coverage of regional dialects and colloquial registers, and consists exclusively of text-based inputs without multimodal content. Finally, as with other large-scale benchmarks based on naturally occurring real-world data, potential overlap with the training corpora of some models cannot be entirely ruled out.

\section{Acknowledgments}
This work was partially supported by the ANR/HELAS chair (ANR-CHIA-0020-01), led by M. Vazirgiannis, which funded members of the authoring group. We also extend our gratitude to C. Stathopoulos for sharing physics teaching Q\&A materials, and to I. Evdaimon and M. Lioudakis for their essential contributions to our initial data collection and processing steps.

\bibliography{aclsmall,googlescholar}

\appendix


\clearpage
\onecolumn
\section{GreekMMLU Tasks and Examples}
\label{example}

\begin{table*}[h]
\centering
\scriptsize
\setlength{\tabcolsep}{5pt}
\begin{tabular}{lllc}
\toprule
\textbf{Task} & \textbf{Tested Concepts} & \textbf{Supercategory} & \textbf{\# Q} \\ \midrule
Accounting & Accounting, balance sheets, microeconomics, institutions... & Social Sciences & 189 \\
Agriculture Professional & Circular bioeconomy, social economy, agri-food systems... & STEM & 344 \\
Agriculture University & Smart agriculture, circular bioeconomy, agri-food systems... & STEM & 75 \\
Art Professional & Applied arts, digital design, materials, fashion
... & Humanities & 605 \\
Art Secondary School & Greek cinema, Greek music, cultural figures, arts history.. & Humanities & 44 \\
Art University & Music theory, rhythm, Greek tradition, ethnomusicology... & Humanities & 17 \\
Biology & Basic biology, microorganisms, human body, animal biology.. & STEM & 423 \\
Chemistry & Thermodynamics, physical chemistry, phase equilibria, surface chemistry...& STEM & 86 \\
Civil Engineering & Building systems, materials, safety, mechanics. & STEM & 774 \\
Clinical Knowledge & Clinical basics, anatomy and physiology, nursing care, dermatology.. & STEM & 638 \\
Computer Networks and Security & Packet-switched networks, architectures, protocol layers, Internet... & STEM & 91 \\
Computer Science Professional & Computing fundamentals, networking, software, digital systems... & STEM & 259 \\
Computer Science University & Computer systems, networking, data analysis, interactive technologies. & STEM & 109 \\
Driving Rules & Traffic regulations, road safety, traffic signs, driver responsibilities. & Other & 1663 \\
Economics Professional & Macroeconomics, international trade, public finance, labor markets. & Social Sciences & 201 \\
Economics University & Strategy, digital business, innovation, technology, costs and pricing.& Social Sciences & 91 \\
Education Professional & Child development, creativity, play-based learning, pedagogy... & Social Sciences & 258 \\
Education University & Educational research, data analysis, academic writing. & Social Sciences & 46 \\
Electrical Engineering & Electrical circuits, sensors, automotive systems, protection devices. & STEM & 557 \\
General Knowledge & Safety procedures, natural phenomena, first aid. & Other & 356 \\
Geography Primary School & Physical and human geography, cartography, regions... & Social Sciences & 365 \\
Geography Secondary School & Physical and human geography, cartography... & Social Sciences & 67 \\
Government and Politics Primary School & Civics, democracy, institutions, governance, rights and duties. & Social Sciences & 285 \\
Government and Politics Secondary School & Political institutions, citizenship, rights, public administration...
 & Social Sciences & 80 \\
Greek History Primary School & Greek history from antiquity to modern times, key events and figures. & Humanities & 469 \\
Greek History Professional & Modern Greek political and constitutional history & Humanities & 122 \\
Greek History Secondary School & Byzantine Empire and Modern Greek State & Humanities & 98 \\
Greek Literature & Authors, Literary Movements and Major Works (Modern and Classical) & Humanities & 19 \\
Greek Mythology & Creation myths, heroic cycles, myth in culture... & Humanities & 243 \\
Greek Traditions & Food Safety, Baking Science and Culinary Operations & Social Sciences & 381 \\
Law & Administrative Law, EU Law and Sports Regulations & Humanities & 941 \\
Management Professional & Management, Economics and Business Operations & Social Sciences & 646 \\
Management University & Human Resource Management & Social Sciences & 30 \\
Mathematics & Mathematical reasoning, algebra, geometry, basic statistics... & STEM & 1123 \\
Medicine Professional & Professional-level medical knowledge and clinical reasoning. & STEM & 467 \\
Medicine University & System-level human physiology and functional integration of organs. & STEM & 77 \\
Maritime Safety and Rescue Operations & Maritime safety, rescue procedures, emergency response at sea.& Other & 153 \\
Modern Greek Language Primary School & Basic grammar, morphology, syntax, orthography.. & Social Sciences & 1483 \\
Modern Greek Language Secondary School & Vocabulary, comprehension, meaning, language use.. & Social Sciences & 885 \\
Physics Primary School & Basic physics, matter, energy, measurements, everyday phenomena. & STEM & 432 \\
Physics Professional & Applied physics and engineering & STEM & 1336 \\
Physics University & Scientific reasoning, physics–biology concepts, thermodynamics, optics... & STEM & 76 \\
Prehistory & Cycladic, Minoan, and Mycenaean civilizations.
 & Humanities & 68 \\
World History & Enlightenment, reformation, renaissance, revolutionary movements... & Humanities & 25 \\
World Religions & Orthodox Christian hymnology. & Humanities & 160 \\ \midrule
\textbf{Total} & & & \textbf{16,857} \\
\bottomrule
\end{tabular}
\caption{Summary of the 45 subjects in the GreekMMLU public dataset. \# Q indicates the total number of questions for each task.}
\label{tab:subject_summary_updated}
\end{table*}

\begin{table*}[h]
 \centering
 \label{tab:greekmmlu_examples}

 \small
 \setlength{\tabcolsep}{3pt}
 \renewcommand{\arraystretch}{1.25}

 \begin{tabular}{|p{2.1cm}|p{7.0cm}|p{6.1cm}|}
 \hline
 \textbf{Subject} & \textbf{Question} & \textbf{Choices} \\
 \hline

 \multirow{2}{*}{STEM} &
 {\selectlanguage{greek}
 Ποια σώματα, όταν δέχονται το ίδιο φως (π.χ. από τον Ήλιο), απορροφούν περισσότερη ενέργεια?
 } &
 {\selectlanguage{greek}
 \underline{{\selectlanguage{greek}Α. Τα σκουρόχρωμα σώματα}} \newline
 Β. Τα ανοιχτόχρωμα σώματα\newline
 Γ. Τα διαφανή σώματα \newline
 Δ. Τα μεταλλικά σώματα
 } \\
 \cline{2-3}

 &
 {\selectlanguage{english}
 Which bodies, when receiving the same light (e.g. from the Sun), absorb more energy?
 } &
 {\selectlanguage{english}
 \underline{A. Dark-colored bodies} \newline
 B. Light-colored bodies \newline
 C. Transparent bodies \newline
 D. Metallic bodies
 } \\
 \hline

 \multirow{2}{*}{Humanities} &
 {\selectlanguage{greek}
 Ποιος σκηνοθέτησε την ταινία «Κυνόδοντας», η οποία υπήρξε το 2011 υποψήφια για Όσκαρ Καλύτερης Ξενόγλωσσης Ταινίας?
 } &
 {\selectlanguage{greek}
 Α. Παντελής Βούλγαρης \newline
 \underline{{\selectlanguage{greek}Β. Γιώργος Λάνθιμος}} \newline
 Γ. Κώστας Γαβράς \newline
 D. Θεόδωρος Αγγελόπουλος
 } \\
 \cline{2-3}

 &
 {\selectlanguage{english}
 Who directed the film Dogtooth, which was nominated in 2011 for the Academy Award for Best Foreign Language Film?
 } &
 {\selectlanguage{english}
 A. Pantelis Voulgaris \newline
 \underline{B. Yorgos Lanthimos }\newline
 C. Costa-Gavras \newline
 D. Theodoros Angelopoulos
 } \\
 \hline

 \multirow{2}{*}{Social Sciences} &
 {\selectlanguage{greek}
 Ποιος είναι ο πρώτος φορέας κοινωνικοποίησης του ατόμου?
 } &
 {\selectlanguage{greek}
 Α. Το σχολείο A \newline
 \underline{{\selectlanguage{greek}Β. Η οικογένεια}} \newline
 Γ. Οι φίλοι/συνομήλικοι \newline
 D. Οι επαγγελματικές σχέσεις
 } \\
 \cline{2-3}

 &
 {\selectlanguage{english}
 What is the primary agent of socialization of an individual?
 } &
 {\selectlanguage{english}
 A. School \newline
 \underline{B. Family} \newline
 C. Friends/peers \newline
 D. Professional relationships
 } \\
 \hline

 \multirow{2}{*}{Other} &
 {\selectlanguage{greek}
 Η Ανώνυμη Εταιρεία (Α.Ε.) είναι εμπορική εταιρεία:
 } &
 {\selectlanguage{greek}
 Α. όταν ενεργεί εμπορικές πράξεις. \newline
 Β. όταν ο σκοπός της είναι εμπορικός. \newline
 \underline{{\selectlanguage{greek}Γ. ανεξάρτητα από τον σκοπό της.}} \newline
 Δ. Η Ανώνυμη Εταιρεία (Α.Ε.) δεν είναι\newline εμπορική εταιρεία.
 } \\
 \cline{2-3}

 &
 {\selectlanguage{english}
 A Société Anonyme (S.A. / public limited company) is considered a commercial company:
 } &
 {\selectlanguage{english}
 A. when it carries out commercial acts. \newline
 B. when its purpose is commercial. \newline
 \underline{C. regardless of its purpose.} \newline
 D. A Société Anonyme is not a commercial company.
 } \\
 \hline

 \multirow{2}{*}{Greek-specific} &
 {\selectlanguage{greek}
 Ποιος ήταν ο Αίολος?
 } &
 {\selectlanguage{greek}
 Α. Ο βασιλιάς των Λαιστρυγόνων \newline
 Β. Ένας σύντροφος του Οδυσσέα \newline
\underline{{\selectlanguage{greek}Γ. Ο θεός των ανέμων}}\newline
 Δ. Ο πατέρας της Κίρκης
 } \\
 \cline{2-3}

 &
 {\selectlanguage{english}
 Who was Aeolus?
 } &
 {\selectlanguage{english}
 A. The king of the Laestrygonians \newline
 B. A companion of Odysseus \newline
 \underline{C. The god of the winds} \newline
 D. The father of Circe
 } \\
 \hline

 \end{tabular}
\caption{Examples from GreekMMLU with their corresponding English translations across different subjects, where the bold items indicate the correct choices.}
\end{table*}

\twocolumn
\section{Statistics of GreekMMLU}
\label{statistic}
\begin{table*}[h]
\centering
\setlength{\tabcolsep}{5pt}
\begin{tabular}{lccccccc}
\toprule
\multirow{2}{*}{Group} &
\multirow{2}{*}{Tasks} &
\multirow{2}{*}{\#Q} &
\multicolumn{3}{c}{Questions per Task} &
\multicolumn{2}{c}{Avg. Tokens} \\
\cmidrule(lr){4-6} \cmidrule(lr){7-8}
& & &
Avg. & Max. & Min. &
Question & Choices \\
\midrule
STEM & 16 & 6787 & 424.19 & 1331 & 70 & 83.91 & 76.86 \\
Humanities & 12 & 2751 & 229.25 & 936 & 12 & 84.29 & 132.65 \\
Social Science & 13 & 4753 & 365.62 & 1478 & 25 & 266.50 & 66.52 \\
Other & 4 & 2341 & 585.25 & 1658 & 148 & 77.83 & 133.96 \\
\midrule
All & 45 & 16632 & 369.60 & 1658 & 12 & 135.30 & 91.17 \\
\bottomrule
\end{tabular}
\caption{Statistics of the GreekMMLU}
\label{tab:greekmmlu_stats}
\end{table*}

 As reported in Table~\ref{tab:greekmmlu_stats}, token-length statistics were obtained using \texttt{tiktoken} with the \texttt{cl100k\_base} encoding. For each question, we computed the number of tokens in the question text (reported as Avg.\ Q Tokens) and the total number of tokens across all answer choices (reported as Avg.\ C Tokens), where choices were concatenated using a single space separator. These values were then averaged across all questions within each subject group to summarize token distribution patterns across the dataset.

\begin{figure*}[h]
\centering
  \includegraphics[width=\linewidth]{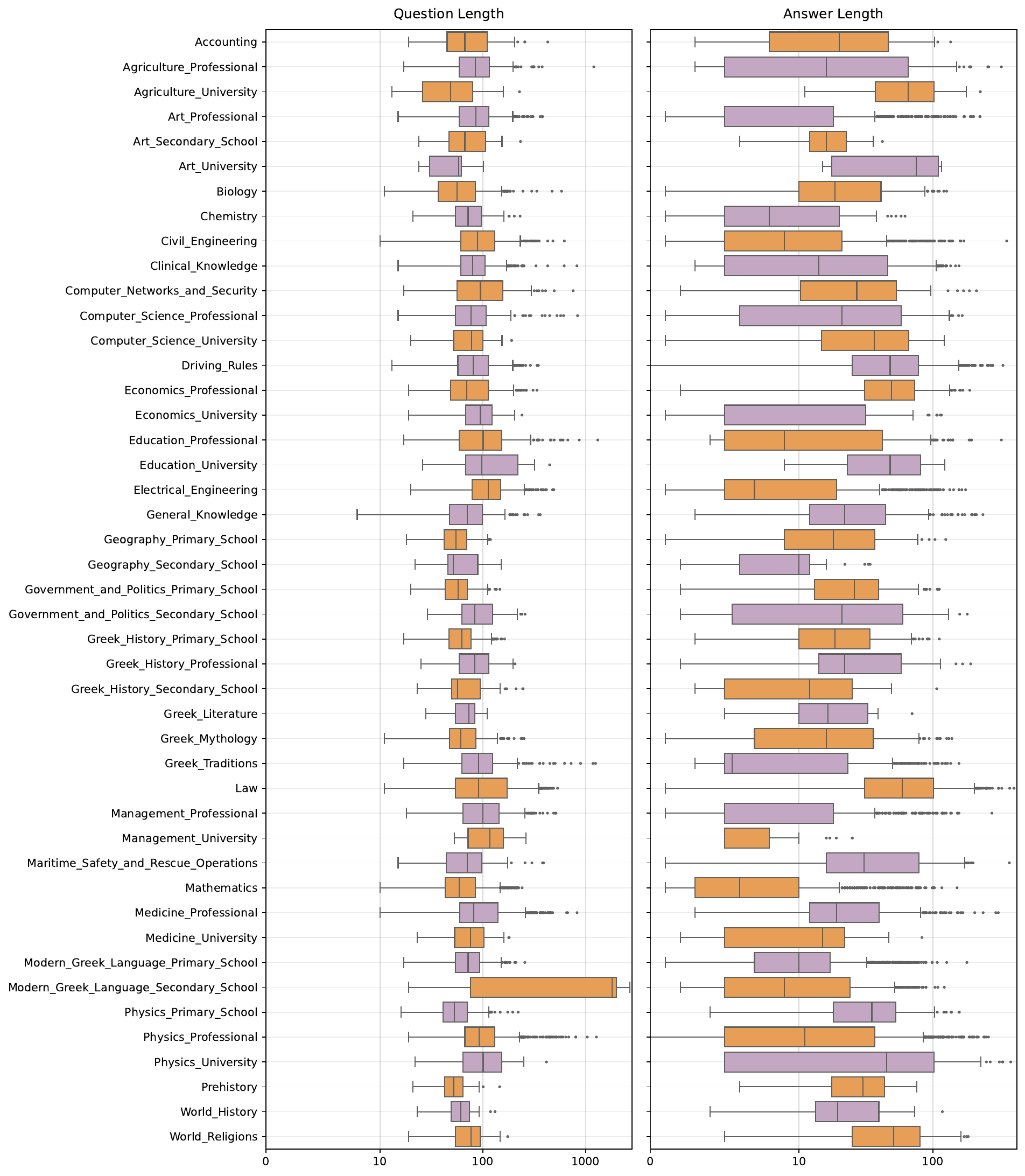}
  \caption{Distribution of character lengths for questions  and answer choices  across all GreekMMLU subjects. }
  \label{fig:subset_length_distribution}
\end{figure*}

As shown in Figure~\ref{fig:subset_length_distribution}, the distributional characteristics of text length vary across subject categories in the GreekMMLU benchmark. The left panel reports question-length distributions, indicating that most subjects exhibit median lengths below 200 characters. Notable exceptions include domains such as Modern Greek Language (Secondary School), which display substantially longer inputs, primarily due to the inclusion of extended reading-comprehension passages. The right panel presents answer-length distributions, which are generally more compact; however, technical and professional domains, including Physics and Law, are characterized by longer answer options, reflecting the increased precision and explanatory detail required in these fields. Overall, this variation in sequence length underscores the benchmark’s ability to evaluate model performance across both short factual queries and longer, context-dependent inputs.

\section{Implementation Details}
\label{sec:implementation_details}

All evaluations were conducted using the lm-evaluation-harness framework \cite{eval-harness} (version 0.4.9.1). The evaluation setup follows a unified configuration derived from the GreekMMLU benchmark, ensuring consistent assessment across all subject domains.


Models were evaluated on the GreekMMLU benchmark that we implemented based on the MMLU format, which consists of subject-specific multiple-choice questions covering STEM, Humanities, Social Sciences, and Other domains. Evaluation was performed under standardized zero-shot and five-shot prompting conditions.


All evaluations for open-source models relied on the default deterministic behavior of the \texttt{lm-evaluation-harness} for multiple-choice tasks. Model predictions were obtained via log-likelihood comparison over answer options rather than generative sampling. 

For closed-source models accessed via external APIs, evaluation was performed using a generation-based multiple-choice protocol. Generation was conducted under deterministic settings, with sampling disabled (\texttt{do\_sample=false}) and temperature set to 0.0.

Model outputs were post-processed using a standardized parsing procedure that extracts the first valid answer symbol, supporting both Greek \selectlanguage{greek}(Α, Β, Γ, Δ)\selectlanguage{english} and Latin (A, B, C, D) representations. When Latin characters were produced, they were deterministically mapped to their Greek equivalents. The extracted prediction was then compared against the answer label to compute accuracy. This procedure supports a variable number of answer choices and was applied uniformly across all evaluated models.




Experiments were conducted on NVIDIA GPUs, primarily using RTX A6000. For larger models exceeding 30B parameters, evaluations were performed on NVIDIA A100 GPUs to accommodate increased memory and compute requirements.

\clearpage
\onecolumn
\section{Experiment results}
\label{experiment_results}

\begin{table*}[h]
\centering
\small
\begin{tabular}{l cccc | c| c}
\toprule
\textbf{Model} & \textbf{STEM} & \textbf{Humanities} & \textbf{Social Sci.} & \textbf{Other} & \textbf{Average} & \textbf{Greek-specific} \\
\midrule
\multicolumn{7}{l}{\textbf{General-Purpose LLMs}} \\
\midrule
Qwen2.5-0.5B & 35.13 & 31.31 & 31.14 & 36.24 & 33.43 & 29.67 \\
Qwen2.5-1.5B & 43.02 & 43.63 & 43.28 & 41.02 & 42.94 & 42.65 \\
Qwen2.5-3B   & 46.22 & 46.23 & 47.97 & 51.62 & 47.46 & 46.09 \\
Qwen2.5-7B   & 62.21 & 62.21 & 64.65 & 61.72 & 62.96 & 65.98 \\
Qwen2.5-14B  & 70.00 & 71.84 & 73.11 & 68.14 & 71.06 & 76.34 \\
Qwen2.5-32B  & 75.60 & 74.40 & 77.52 & 72.72 & 75.73 & 79.21 \\
Qwen2.5-72B  & 82.22 & 82.84 & 83.19 & 78.60 & 82.18 & 85.79 \\
Qwen3-0.6B   & 41.74 & 38.43 & 40.51 & 42.21 & 40.95 & 38.33 \\
Qwen3-1.7B   & 59.61 & 52.99 & 58.21 & 57.14 & 57.97 & 57.46 \\
Qwen3-4B     & 69.34 & 60.79 & 67.93 & 64.51 & 67.14 & 67.95 \\
Qwen3-8B     & 74.05 & 68.78 & 73.91 & 69.69 & 72.77 & 75.85 \\
Qwen3-14B    & 78.27 & 72.20 & 79.53 & 73.17 & 77.26 & 81.17 \\
Qwen3-30B    & 81.74 & 75.31 & 80.46 & 76.80 & 79.86 & 82.32 \\

\addlinespace[0.5em]
Llama-2-7b-hf & 36.53 & 35.60 & 33.53 & 33.15 & 34.99 & 32.87 \\
Llama-3-8B   & 61.79 & 65.59 & 66.76 & 64.11 & 64.24 & 68.96 \\
Llama-3.1-70B& 77.62 & 82.79 & 83.66 & 78.35 & 80.41 & 87.30 \\
Llama-3.1-8B & 61.15 & 63.58 & 65.89 & 62.77 & 63.25 & 67.21 \\
Llama-3.2-1B & 34.54 & 33.73 & 34.26 & 37.08 & 34.65 & 32.68 \\
Llama-3.2-3B & 46.91 & 49.93 & 48.88 & 50.57 & 48.42 & 46.53 \\

\addlinespace[0.5em]
Gemma-3-1B-pt  & 30.94 & 30.72 & 30.23 & 32.16 & 30.82 & 30.03 \\
Gemma-3-4B-pt  & 62.53 & 64.95 & 67.31 & 63.27 & 64.54 & 69.75 \\
Gemma-3-12B-pt & 77.16 & 78.59 & 77.48 & 76.16 & 77.34 & 79.48 \\
Gemma-3-27B-pt & 82.29 & 82.88 & 83.68 & 80.69 & 82.64 & 86.53 \\

\addlinespace[0.5em]
XGLM-1.7B & 32.57 & 38.53 & 29.55 & 26.50 & 31.71 & 30.26 \\
XGLM-2.9B & 30.79 & 41.85 & 28.34 & 29.34 & 30.72 & 29.66 \\
XGLM-4.5B & 33.44 & 41.02 & 29.45 & 27.92 & 32.37 & 31.56 \\
XGLM-7.5B & 30.79 & 41.85 & 28.34 & 29.34 & 30.72 & 29.66 \\
GLM-4-9B  & 67.70 & 69.69 & 74.00 & 68.79 & 70.20 & 74.23 \\

\midrule
\multicolumn{7}{l}{\textbf{Greek and European LLMs}} \\
\midrule
Llama-Krikri-8B-Base & 66.75 & 75.63 & 75.11 & 68.09 & 70.88 & 80.22 \\
Meltemi-7B-v1       & 60.26 & 64.58 & 51.19 & 64.91 & 58.38 & 49.10 \\
Meltemi-7B-v1.5     & 59.88 & 67.14 & 64.11 & 65.95 & 62.99 & 67.10 \\

\addlinespace[0.5em]
EuroLLM-1.7B & 32.71 & 30.94 & 30.71 & 36.54 & 32.27 & 30.49 \\
EuroLLM-9B   & 60.91 & 69.92 & 68.00 & 66.75 & 65.18 & 73.03 \\
EuroLLM-22B  & 71.65 & 77.23 & 77.01 & 71.13 & 74.11 & 82.32 \\

\midrule
Random Baseline & 32.33 & 28.77 & 31.86 & 32.62 & 30.42 & 31.59 \\
\bottomrule
\end{tabular}
\caption{Overall five-shot performance of \textbf{base} LLMs on the GreekMMLU benchmark. Accuracy (\%) is reported. The \textit{Greek-specific} column includes an average of History, Traditions, and Mythology subsets.}
\label{tab:5s_base_main_results}
\end{table*}

\begin{table*}[h]
\centering
\small
\begin{tabular}{l cccc | c| c}
\toprule
\textbf{Model} & \textbf{STEM} & \textbf{Humanities} & \textbf{Social Sci.} & \textbf{Other} & \textbf{Average} & \textbf{Greek-specific} \\
\midrule
\multicolumn{7}{l}{\textbf{General-Purpose LLMs}} \\
\midrule
GPT-5.2                  & 88.29 & 90.28 & 90.74 & 86.76 & 89.18 & 93.85 \\
GPT-4o                   & 85.62 & 90.05 & 90.29 & 86.16 & 87.83 & 93.22 \\
Gemini 3 Flash           & 91.24 & 88.63 & 98.10 & 87.46 & 91.20 & 93.52 \\
\addlinespace[0.5em]
Qwen2.5-0.5B-Instruct    & 34.82 & 32.72 & 30.58 & 36.98 & 33.39 & 29.26 \\
Qwen2.5-1.5B-Instruct    & 46.66 & 45.82 & 46.86 & 45.84 & 46.52 & 45.98 \\
Qwen2.5-3B-Instruct      & 51.82 & 52.67 & 54.58 & 56.20 & 53.39 & 55.11 \\
Qwen2.5-7B-Instruct      & 64.20 & 63.35 & 64.85 & 63.36 & 64.20 & 67.27 \\
Qwen2.5-14B-Instruct     & 69.90 & 70.52 & 72.20 & 68.59 & 70.59 & 75.36 \\
Qwen2.5-32B-Instruct     & 75.69 & 73.66 & 78.72 & 72.22 & 76.01 & 81.39 \\
Qwen2.5-72B-Instruct     & 80.93 & 81.65 & 83.14 & 78.25 & 81.44 & 85.90 \\
Qwen3-4B-Instruct-2507   & 70.46 & 62.25 & 68.66 & 66.80 & 68.32 & 69.59 \\
Qwen3-30B-Instruct       & 82.42 & 77.09 & 81.50 & 77.90 & 80.85 & 83.42 \\
\addlinespace[0.5em]
Llama-2-7b-chat-hf       & 38.03 & 39.66 & 35.58 & 33.10 & 36.83 & 35.60 \\
Llama-3-8B-Instruct      & 61.91 & 65.40 & 68.55 & 64.31 & 64.88 & 71.26 \\
Llama-3.1-8B-Instruct    & 61.77 & 67.09 & 68.22 & 62.72 & 64.74 & 71.12 \\
Llama-3.1-70B-Instruct   & 79.09 & 81.93 & 83.34 & 77.90 & 80.74 & 86.80 \\
Llama-3.2-1B-Instruct    & 40.93 & 40.58 & 38.98 & 43.01 & 40.49 & 36.15 \\
Llama-3.2-3B-Instruct    & 48.36 & 49.34 & 52.19 & 54.06 & 50.46 & 51.99 \\
Llama-3.3-70B-Instruct   & 79.65 & 82.47 & 84.19 & 79.09 & 81.47 & 87.90 \\
\addlinespace[0.5em]
Mistral-7B-Instruct-v0.3 & 50.10 & 51.94 & 52.72 & 53.11 & 51.58 & 53.52 \\
\addlinespace[0.5em]
Gemma-3-1B-it            & 45.71 & 46.33 & 45.01 & 46.54 & 45.66 & 43.80 \\
Gemma-3-4B-it            & 62.84 & 63.62 & 68.16 & 63.27 & 64.77 & 70.00 \\
Gemma-3-12B-it           & 74.51 & 75.35 & 79.79 & 74.27 & 76.35 & 81.97 \\
Gemma-3-27B-it           & 80.51 & 81.70 & 83.05 & 78.50 & 81.27 & 86.12 \\
\addlinespace[0.5em]
Aya-101 & 54.37 & 54.71 & 64.78 & 57.94 & 58.59 & 62.62 \\
Aya-expanse-8b           & 62.18 & 66.41 & 69.80 & 65.11 & 65.64 & 71.07 \\

BLOOMZ-1b1 & 32.17 & 39.48 & 29.11 & 31.62 & 31.60 & 30.02 \\
BLOOMZ-1b7 & 31.78 & 39.24 & 28.18 & 29.06 & 30.95 & 29.13 \\
BLOOMZ-7b1 & 31.88 & 32.04 & 30.02 & 30.86 & 31.16 & 28.88\\
mT0-large & 31.56 & 40.31 & 27.97 & 30.20 & 30.88 & 28.84 \\
mT0-xl & 39.54 & 47.44 & 41.08 & 46.44 & 41.00 & 41.34 \\
mT0-xxl & 48.95 & 43.72 & 55.58 & 53.69 & 51.14 & 55.36 \\
GLM-4-9B-chat            & 67.31 & 66.54 & 72.15 & 67.40 & 68.83 & 72.60 \\
\midrule
\multicolumn{7}{l}{\textbf{Greek and European LLMs}} \\
\midrule
Llama-Krikri-8B-Instruct & 67.04 & 75.58 & 74.57 & 67.94 & 70.80 & 79.21 \\
Meltemi-7B-Instruct-v1.5 & 58.91 & 66.86 & 68.24 & 64.11 & 63.71 & 69.97 \\
Plutus-8B-instruct       & 67.23 & 74.81 & 74.42 & 68.34 & 70.77 & 78.36 \\
\addlinespace[0.5em]
EuroLLM-1.7B-Instruct    & 32.87 & 31.36 & 30.73 & 36.29 & 32.37 & 31.07 \\
EuroLLM-9B-Instruct      & 62.37 & 72.43 & 70.88 & 67.75 & 67.20 & 75.96 \\
EuroLLM-22B-Instruct-2512 & 72.09 & 78.73 & 78.01 & 72.92 & 75.05 & 83.06 \\
\midrule
Random Baseline          & 32.33 & 28.77 & 31.86 & 32.62 & 30.42 & 31.59 \\
\bottomrule
\end{tabular}
\caption{Overall five-shot performance of \textbf{instruction-tuned} LLMs on the GreekMMLU benchmark. Accuracy (\%) is reported. The \textit{Greek-specific} column includes an average of History, Traditions, and Mythology subsets.}
\label{tab:5s_it_main_results}
\end{table*}

\begin{table*}[h]
\centering
\small
\begin{tabular}{l cccc | c| c}
\toprule
\textbf{Model} & \textbf{STEM} & \textbf{Humanities} & \textbf{Social Sci.} & \textbf{Other} & \textbf{Average} & \textbf{Greek-specific} \\
\midrule
\multicolumn{7}{l}{\textbf{General-Purpose LLMs}} \\
\midrule
Qwen2.5-0.5B                 & 35.20 & 33.32 & 34.18 & 35.74 & 34.68 & 32.30 \\
Qwen2.5-1.5B                 & 36.19 & 35.33 & 34.27 & 39.62 & 35.85 & 31.48 \\
Qwen2.5-3B                   & 46.34 & 44.64 & 45.02 & 40.32 & 44.94 & 44.86 \\
Qwen2.5-7B                   & 51.92 & 56.37 & 59.34 & 53.31 & 55.16 & 61.07 \\
Qwen2.5-14B                  & 64.11 & 64.95 & 66.04 & 60.23 & 64.39 & 67.95 \\
Qwen2.5-32B                  & 72.74 & 71.29 & 75.26 & 70.68 & 73.14 & 77.40 \\
Qwen2.5-72B                  & 78.31 & 78.78 & 81.35 & 76.75 & 79.20 & 83.91 \\
Qwen3-0.6B                   & 36.35 & 34.50 & 34.95 & 36.64 & 35.67 & 33.11 \\
Qwen3-1.7B                   & 49.62 & 44.00 & 49.70 & 49.23 & 48.85 & 47.87 \\
Qwen3-4B                     & 64.92 & 57.78 & 64.67 & 61.27 & 63.44 & 64.97 \\
Qwen3-8B                     & 70.27 & 63.49 & 69.89 & 66.50 & 68.78 & 71.17 \\
Qwen3-14B                    & 62.59 & 61.89 & 69.11 & 67.89 & 65.32 & 68.20 \\
Qwen3-30B                    & 70.72 & 69.06 & 74.95 & 59.63 & 70.56 & 77.49 \\

\addlinespace[0.5em]
Llama-2-7b-hf                & 36.23 & 35.97 & 33.80 & 35.54 & 35.30 & 32.84 \\
Llama-3-8B                   & 52.39 & 57.01 & 56.54 & 50.02 & 54.10 & 56.75 \\
Llama-3.1-8B                 & 51.08 & 57.42 & 54.78 & 50.62 & 53.10 & 54.78 \\
Llama-3.1-70B                & 72.42 & 79.05 & 78.90 & 74.46 & 75.71 & 82.46 \\
Llama-3.2-1B                 & 37.37 & 36.51 & 34.86 & 36.83 & 36.35 & 33.66 \\
Llama-3.2-3B                 & 41.82 & 41.67 & 43.81 & 44.75 & 42.82 & 41.97 \\
\addlinespace[0.5em]
Gemma-3-1B-pt                & 33.11 & 32.72 & 30.29 & 31.41 & 31.91 & 30.00 \\
Gemma-3-4B-pt                & 50.77 & 52.67 & 53.83 & 52.17 & 52.21 & 55.00 \\
Gemma-3-12B-pt               & 73.30 & 74.44 & 78.28 & 72.27 & 74.99 & 80.71 \\
Gemma-3-27B-pt               & 77.81 & 79.83 & 80.52 & 76.95 & 78.88 & 83.33 \\
\addlinespace[0.5em]

XGLM-1.7B                    & 34.07 & 35.01 & 31.55 & 27.35 & 32.98 & 31.67 \\
XGLM-2.9B                    & 33.25 & 29.39 & 34.94 & 26.92 & 32.68 & 32.90 \\
XGLM-4.5B                    & 33.93 & 36.86 & 31.26 & 27.07 & 33.00 & 31.42 \\
XGLM-7.5B                    & 33.97 & 34.72 & 29.86 & 24.79 & 32.16 & 30.11 \\
GLM-4-9B                     & 63.19 & 63.81 & 68.22 & 63.41 & 64.98 & 68.52 \\
\midrule
\multicolumn{7}{l}{\textbf{Greek and European LLMs}} \\
\midrule
Llama-Krikri-8B-Base         & 59.83 & 68.92 & 65.56 & 62.92 & 63.33 & 68.72 \\
Meltemi-7B-v1                & 52.90 & 56.64 & 43.71 & 56.45 & 50.76 & 40.98 \\
Meltemi-7B-v1.5              & 52.13 & 55.23 & 53.74 & 52.36 & 53.11 & 56.28 \\
\addlinespace[0.5em]
EuroLLM-1.7B                 & 33.21 & 34.28 & 30.54 & 32.11 & 32.33 & 29.86 \\
EuroLLM-9B                   & 62.52 & 71.11 & 69.46 & 65.21 & 66.30 & 73.74 \\
EuroLLM-22B                  & 66.94 & 75.35 & 73.49 & 68.44 & 70.43 & 77.46 \\

\midrule
Random Baseline               & 32.33 & 28.77 & 31.86 & 32.62 & 30.42 & 31.59 \\
\bottomrule
\end{tabular}
\caption{Overall zero-shot performance of \textbf{base} LLMs on the GreekMMLU benchmark. Accuracy (\%) is reported. The \textit{Greek-specific} column includes an average of History, Traditions, and Mythology subsets.}
\label{tab:0s_base_main_results}
\end{table*}

\begin{table*}[h]
\centering
\small
\begin{tabular}{l cccc | c| c}
\toprule
\textbf{Model} & \textbf{STEM} & \textbf{Humanities} & \textbf{Social Sci.} & \textbf{Other} & \textbf{Average} & \textbf{Greek-specific} \\
\midrule
\multicolumn{7}{l}{\textbf{General-Purpose LLMs}} \\
\midrule
GPT-5.2                         & 86.05 & 88.27 & 90.29 & 85.96 & 87.75 & 92.92 \\
GPT-4o                          & 84.54 & 88.68 & 89.36 & 85.42 & 86.81 & 93.11 \\
Gemini 3 Flash                  & 92.82 & 92.88 & 94.16 & 91.84 & 93.16 & 95.44 \\
\addlinespace[0.5em]
Qwen2.5-0.5B-Instruct        & 35.20 & 33.91 & 34.40 & 36.24 & 34.89 & 32.51 \\
Qwen2.5-1.5B-Instruct        & 34.08 & 34.19 & 33.33 & 34.69 & 33.92 & 34.86 \\
Qwen2.5-3B-Instruct          & 50.67 & 50.07 & 51.26 & 48.28 & 50.50 & 51.97 \\
Qwen2.5-7B-Instruct          & 59.88 & 58.33 & 61.98 & 58.89 & 60.25 & 64.02 \\
Qwen2.5-14B-Instruct         & 65.20 & 66.41 & 69.78 & 62.92 & 66.61 & 73.06 \\
Qwen2.5-32B-Instruct         & 72.08 & 71.47 & 76.61 & 69.69 & 73.22 & 80.03 \\
Qwen2.5-72B-Instruct         & 78.90 & 79.14 & 81.92 & 76.95 & 79.70 & 84.67 \\
Qwen3-4B-Instruct-2507       & 65.35 & 57.60 & 66.64 & 63.46 & 64.52 & 67.16 \\
Qwen3-30B-Instruct           & 79.31 & 74.81 & 79.33 & 76.56 & 78.39 & 81.80 \\
\addlinespace[0.5em]
Llama-2-7b-chat-hf           & 36.63 & 34.92 & 34.26 & 33.85 & 35.27 & 33.61 \\
Llama-3-8B-Instruct     & 59.01 & 62.85 & 64.16 & 60.38 & 61.40 & 65.38 \\
Llama-3.1-8B-Instruct    & 56.58 & 62.85 & 62.56 & 57.84 & 59.56 & 64.75 \\
Llama-3.1-70B-Instruct   & 76.35 & 82.34 & 82.57 & 75.61 & 79.13 & 86.99 \\
Llama-3.2-1B-Instruct        & 38.01 & 37.33 & 36.87 & 35.94 & 37.29 & 35.46 \\
Llama-3.2-3B-Instruct        & 43.77 & 43.04 & 45.62 & 45.64 & 44.52 & 46.15 \\
Llama-3.3-70B-Instruct   & 77.03 & 82.20 & 82.92 & 76.80 & 79.65 & 86.94 \\
\addlinespace[0.5em]
Mistral-7B-Instruct-v0.3     & 48.05 & 48.93 & 51.23 & 48.23 & 49.25 & 52.02 \\
Mistral-Small-24B-Instruct-2501 & 67.16 & 71.72 & 74.79 & 66.39 & 70.47 & 78.06 \\
\addlinespace[0.5em]
Gemma-3-1B-it                & 45.25 & 44.45 & 44.19 & 46.59 & 44.95 & 42.05 \\
Gemma-3-4B-it                & 59.79 & 60.43 & 65.95 & 62.32 & 62.24 & 68.58 \\
Gemma-3-12B-it               & 72.95 & 75.13 & 79.28 & 72.62 & 75.31 & 82.21 \\
Gemma-3-27B-it               & 78.03 & 79.87 & 82.19 & 75.96 & 79.41 & 85.33 \\
\addlinespace[0.5em]
Aya-101 & 52.62 & 53.75 & 62.16 & 57.05 & 56.73 & 59.86 \\
Aya-expanse-8b               & 60.75 & 65.13 & 68.33 & 63.27 & 64.17 & 69.51 \\

BLOOMZ-1b1 & 35.64 & 34.37 & 34.26 & 36.14 & 35.07 & 32.68 \\
BLOOMZ-1b7 & 35.66 & 34.19 & 33.15 & 35.94 & 34.66 & 31.53 \\
BLOOMZ-7b1 & 34.27 & 32.63 & 30.03 & 32.30 & 32.40 & 28.80 \\
mT0-large & 36.89 & 36.47 & 37.09 & 37.28 & 36.95 & 34.95 \\
mT0-xl & 46.44 & 47.88 & 54.21 & 52.51 & 49.96 & 52.24 \\
mT0-xxl & 52.72 & 53.13 & 61.33 & 36.92 & 56.57 & 56.91 \\
GLM-4-9B-chat                & 61.44 & 64.26 & 68.42 & 65.85 & 64.68 & 69.29 \\
\midrule
\multicolumn{7}{l}{\textbf{Greek and European LLMs}}  \\
\midrule
Llama-Krikri-8B-Instruct     & 62.62 & 70.29 & 70.57 & 64.11 & 66.47 & 74.73 \\
Meltemi-7B-Instruct-v1.5     & 57.18 & 63.99 & 64.31 & 61.03 & 60.93 & 66.42 \\
Plutus-8B-instruct           & 61.65 & 69.74 & 69.73 & 64.01 & 65.71 & 73.96 \\
\addlinespace[0.5em]
EuroLLM-1.7B-Instruct        & 29.47 & 30.76 & 28.76 & 31.76 & 29.68 & 30.16 \\
EuroLLM-9B-Instruct          & 64.42 & 73.16 & 72.24 & 66.85 & 68.48 & 76.86 \\
EuroLLM-22B-Instruct-2512    & 69.78 & 75.31 & 74.66 & 70.08 & 72.18 & 78.99 \\
\midrule
Random Baseline           & 32.33 & 28.77 & 31.86 & 32.62 & 30.42 & 31.59 \\
\bottomrule
\end{tabular}
\caption{Overall zero-shot performance of \textbf{instruction-tuned} LLMs on the GreekMMLU benchmark. Accuracy (\%) is reported. The \textit{Greek-specific} column includes an average of History, Traditions, and Mythology subsets.}
\label{tab:0s_it_main_results}
\end{table*}

\clearpage
\twocolumn
\section{Comparison Between the Public and Private Results}
\label{corrpubpri}
\begin{table}[h]
    \centering
    \begin{tabular}{lcc}
        \toprule
        \textbf{Category} & \textbf{Pearson $r$} & \textbf{$p$-value} \\
        \midrule
        STEM & 0.9973 & $3.18 \times 10^{-74}$ \\
        Humanities & 0.9895 & $1.98 \times 10^{-55}$ \\
        Social Sci. & 0.9966 & $5.01 \times 10^{-71}$ \\
        Other & 0.9929 & $7.42 \times 10^{-61}$ \\
        Average & 0.9984 & $1.85 \times 10^{-81}$ \\
        \midrule
        Combined & 0.9908 & $3.27 \times 10^{-287}$ \\
        \bottomrule
    \end{tabular}
    \caption{Correlation between Public and Private GreekMMLU Scores (5-shot).}
    \label{tab:greekmmlu_correlation5}
\end{table}

\begin{table}[h]
    \centering
    \begin{tabular}{lcc}
        \toprule
        \textbf{Category} & \textbf{Pearson $r$} & \textbf{$p$-value} \\
        \midrule
        STEM & 0.9986 & $2.83 \times 10^{-83}$ \\
        Humanities & 0.9934 & $5.39 \times 10^{-62}$ \\
        Social Sci. & 0.9970 & $5.96 \times 10^{-73}$ \\
        Other & 0.9881 & $1.01 \times 10^{-53}$ \\
        Average & 0.9988 & $4.17 \times 10^{-86}$ \\
        \midrule
        Combined & 0.9901 & $4.74 \times 10^{-282}$ \\
        \bottomrule
    \end{tabular}
    \caption{Correlation between Public and Private GreekMMLU Scores (0-shot).}
    \label{tab:greekmmlu_correlation0}
\end{table}

\begin{figure}[t]
    \centering
  \includegraphics[width=0.8\columnwidth]{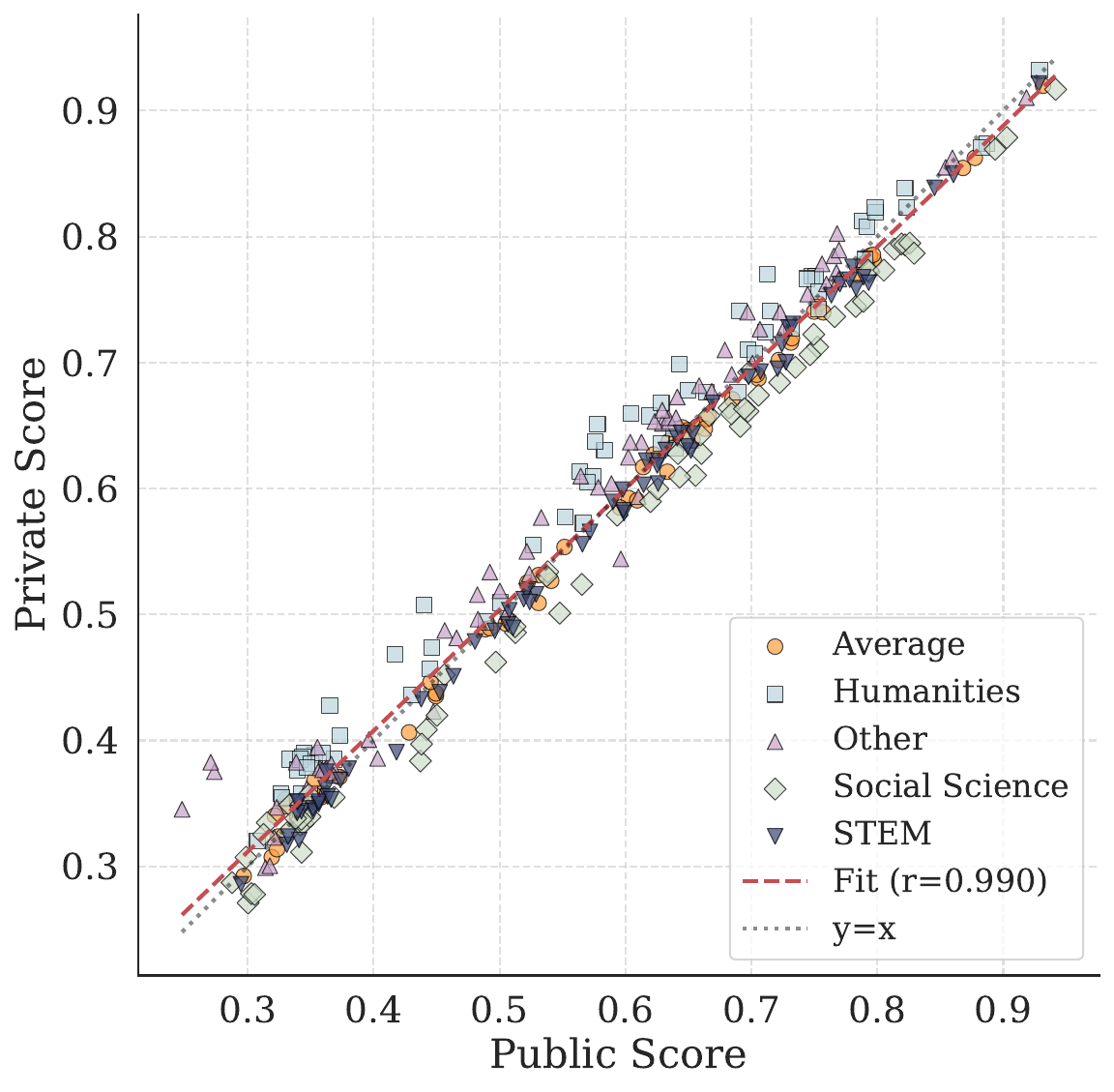}
  \caption{Correlation between Public and Private GreekMMLU Scores
(0-shot)}
  \label{fig:corr0}
\end{figure}

\begin{figure}[t]
    \centering
  \includegraphics[width=0.8\columnwidth]{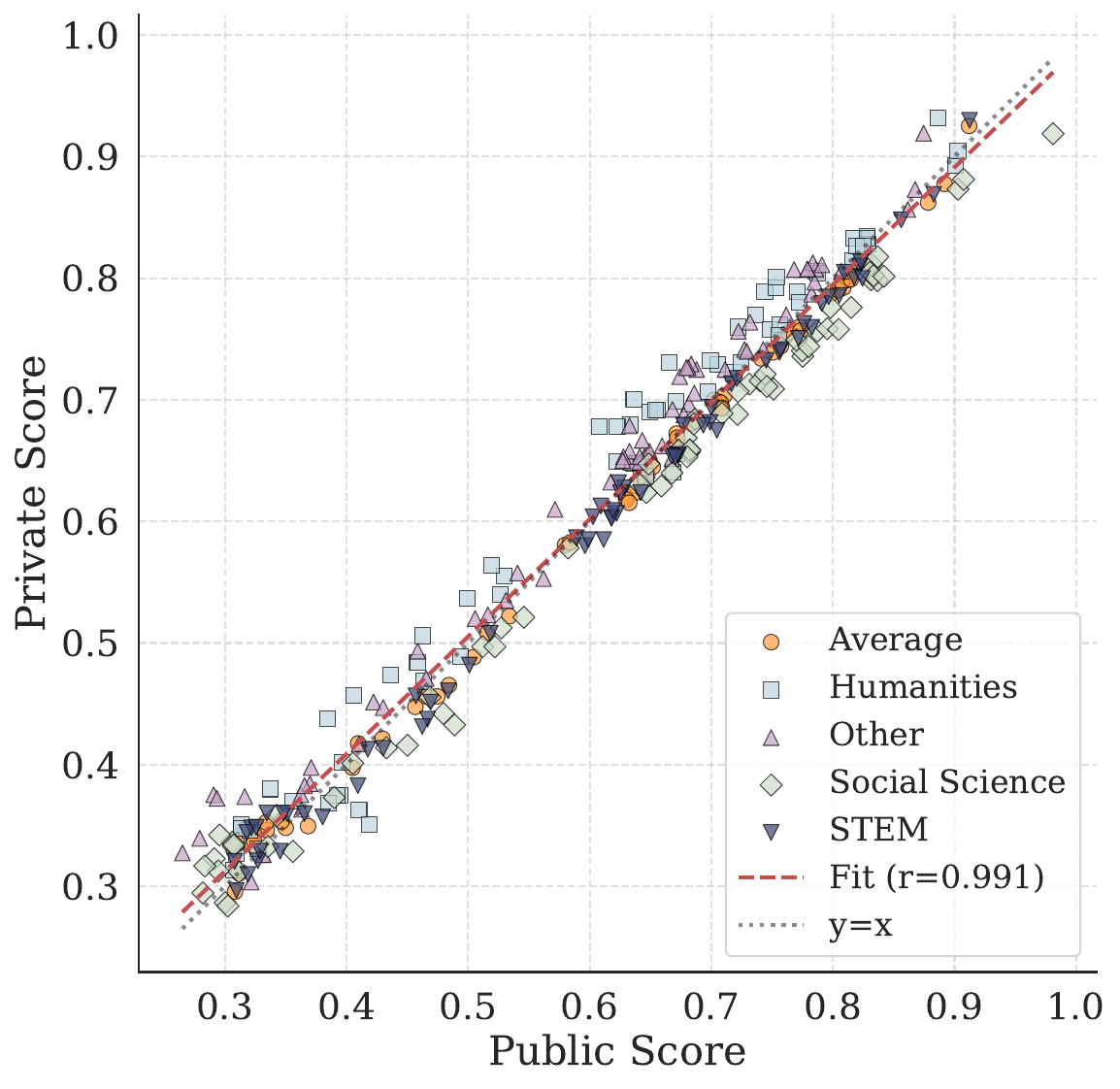}
  \caption{Correlation between Public and Private GreekMMLU Scores
(5-shot)}
  \label{fig:corr5}
\end{figure}

To verify that the private split of GreekMMLU provides a reliable estimate of model performance, we analyze the correlation between public and private evaluation results for each model. Figures~\ref{fig:corr0} and~\ref{fig:corr5} show a strong linear relationship between public and private scores under zero-shot and five-shot settings, respectively. This observation is quantified in Tables~\ref{tab:greekmmlu_correlation0} and~\ref{tab:greekmmlu_correlation5}, where Pearson correlation coefficients exceed 0.98 across all subject categories and evaluation setups, with extremely small $p$-values.

The consistently high correlations across STEM, Humanities, Social Sciences, and Other domains indicate that relative model rankings are well preserved between the two splits. These results demonstrate that the private GreekMMLU set faithfully reflects model performance observed on the public data, supporting its use as a robust and contamination-resistant benchmark for leaderboard-based evaluation.

\section{Extended Calibration}
\label{calib}
We evaluated the calibration of a wider set of 11 models in the 5-shot setting (Table~\ref{tab:calibration_extended}). The results highlight three distinct behaviors:
\begin{table}[h]
    \centering
    \begin{tabular}{lc}
        \toprule
        \textbf{Model} & \textbf{Pearson $r$} \\
        \midrule
        Llama-Krikri-8B-Base & 0.96 \\
        EuroLLM-22B & 0.95 \\
        Llama-3.1-70B & 0.95 \\
        Llama-3.1-8B & 0.95 \\
        Qwen3-30B & 0.93 \\
        EuroLLM-9B & 0.93 \\
        Qwen2.5-72B & 0.92 \\
        Llama-3.3-70B & 0.91 \\
        Qwen2.5-32B & 0.88 \\
        Qwen2.5-7B & 0.87 \\
        Llama-2-7b-hf & 0.46 \\
        \bottomrule
    \end{tabular}
    \caption{Extended calibration analysis ranked by Pearson correlation coefficient ($r$) in the 5-shot setting.}
    \label{tab:calibration_extended}
\end{table}

We evaluated the calibration of a wider set of 11 models in the 5-shot setting (Table~\ref{tab:calibration_extended}). The results highlight three distinct behaviors. The top-performing models, including the Greek-specific {Llama-Krikri-8B} and the {Llama 3.1} family, fall into a {highly calibrated} cluster, exhibiting the highest correlation ($r \geq 0.95$) and accurately reflecting their true probability of correctness. A second group, comprising the {Qwen 2.5/3} family, shows {moderate calibration} with strong but slightly lower correlation coefficients ($r \approx 0.87-0.93$). Finally, the older {Llama-2-7b} baseline remains {poorly calibrated} with significantly lower correlation ($r=0.46$)

\section{Subject-Level Performance Distribution} \label{sec:subject_distribution}
\begin{figure*}[h]
\centering
\includegraphics[width=\textwidth]{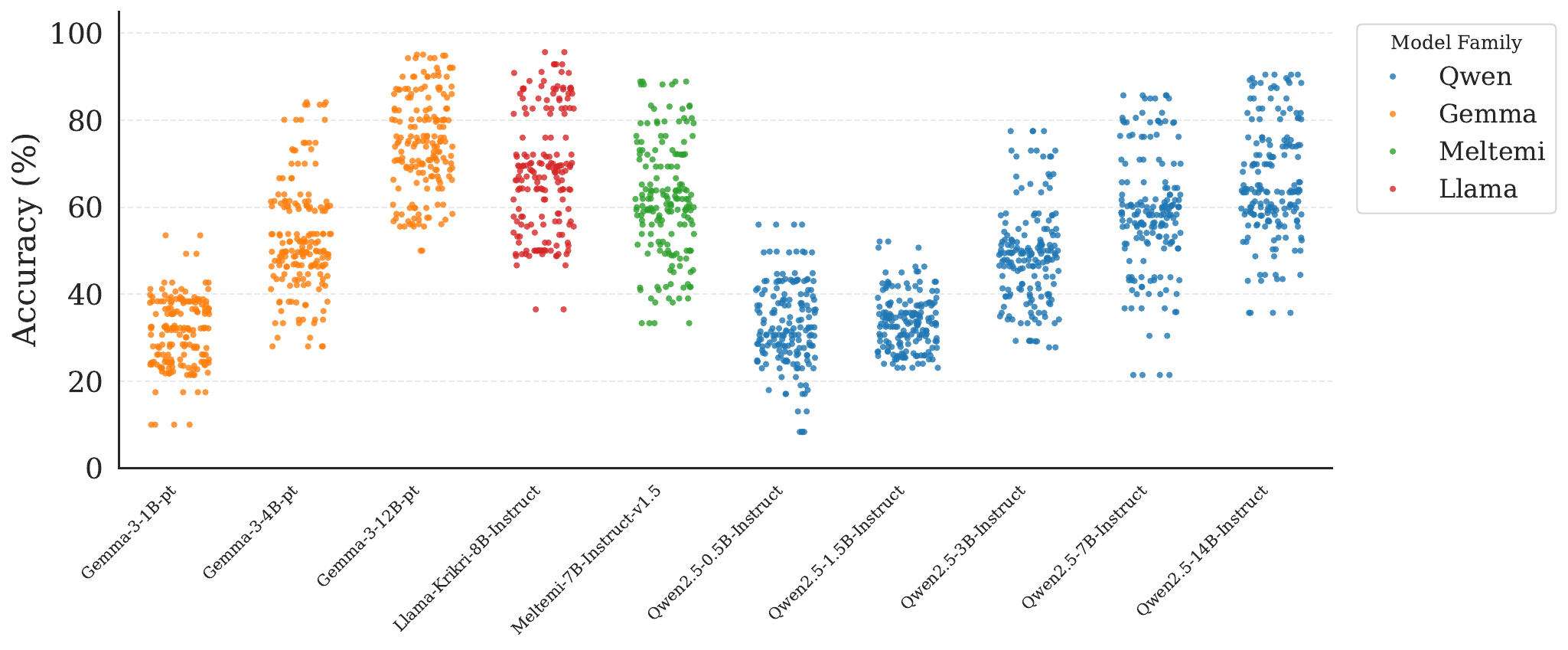}
\caption{Subject performance distribution (0-shot)}
\label{fig:subject_distribution_0shot}
\end{figure*}

\begin{figure*}[h]
    \centering
    \includegraphics[width=\textwidth]{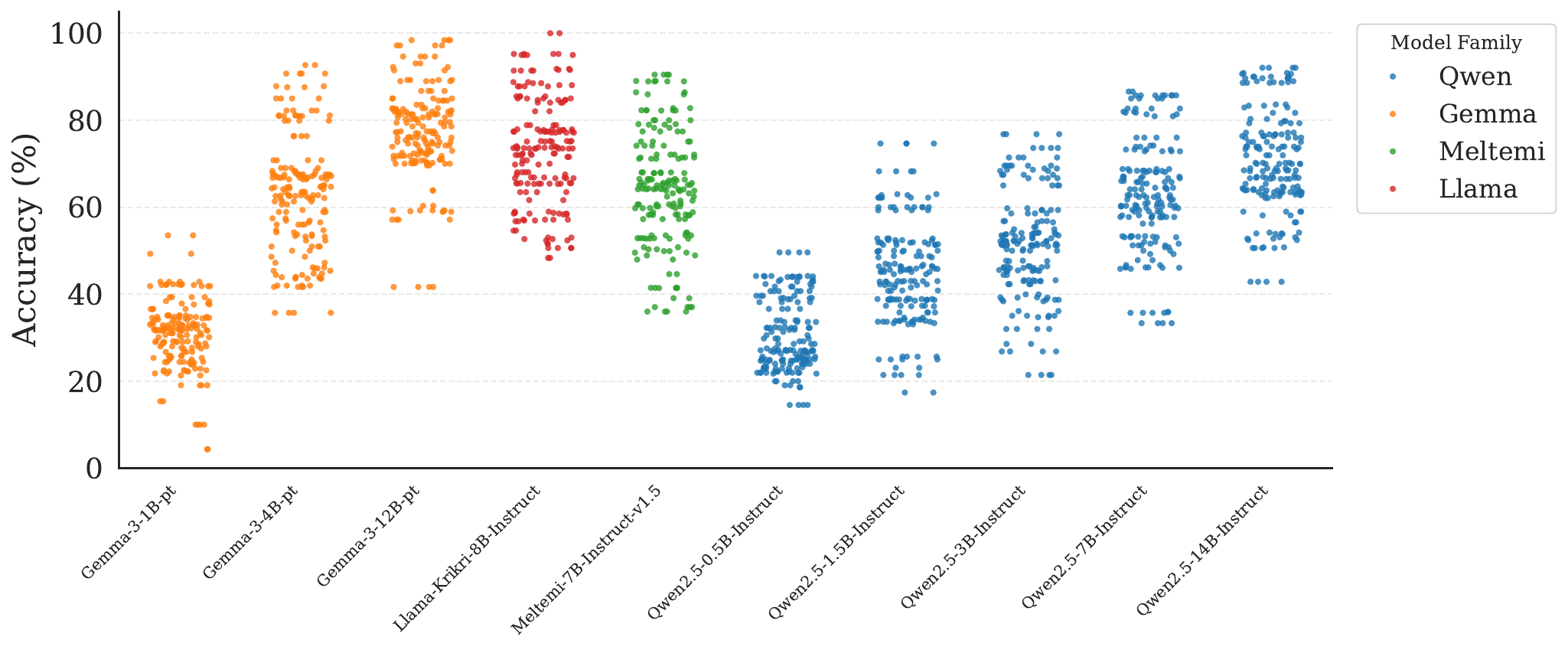}
    \caption{Subject performance distribution (5-shot)}
    \label{fig:subject_distribution_5shot}
\end{figure*}

Figures~\ref{fig:subject_distribution_0shot} and \ref{fig:subject_distribution_5shot} illustrate the distribution of accuracy scores across the 45 subjects of the GreekMMLU benchmark in zero-shot and five-shot settings, respectively, offering granular insights into model robustness beyond aggregate metrics. The analysis consistently reveals a distinct relationship between model specialization, scale, and cross-domain consistency across both prompting strategies. Notably, the Greek-centric \texttt{Llama-Krikri-8B} demonstrates a significant upward shift in its performance distribution relative to generic models of comparable size, such as \texttt{Llama-3.1-8B} and \texttt{EuroLLM-9B}. Its elevated performance baseline indicates a resilience against catastrophic failure modes on linguistically or culturally complex subjects, whereas generic counterparts frequently degrade to near-random accuracy in these areas.

Furthermore, the data highlights the stabilizing effect of model scale. The largest evaluated model, \texttt{Llama-3.1-70B}, exhibits the tightest clustering of subject scores within the upper quartile in both settings. This suggests that massive parameterization functions as a stabilizing factor, ensuring consistent competency across niche domains where smaller architectures struggle. In contrast, smaller models like \texttt{Qwen2.5-0.5B} display extreme vertical variance; while they occasionally achieve parity on simpler tasks, they lack the generalization capabilities required to maintain robust performance across the full breadth of the academic curriculum.

\begin{figure*}[h]
    \centering
    \includegraphics[width=\textwidth]{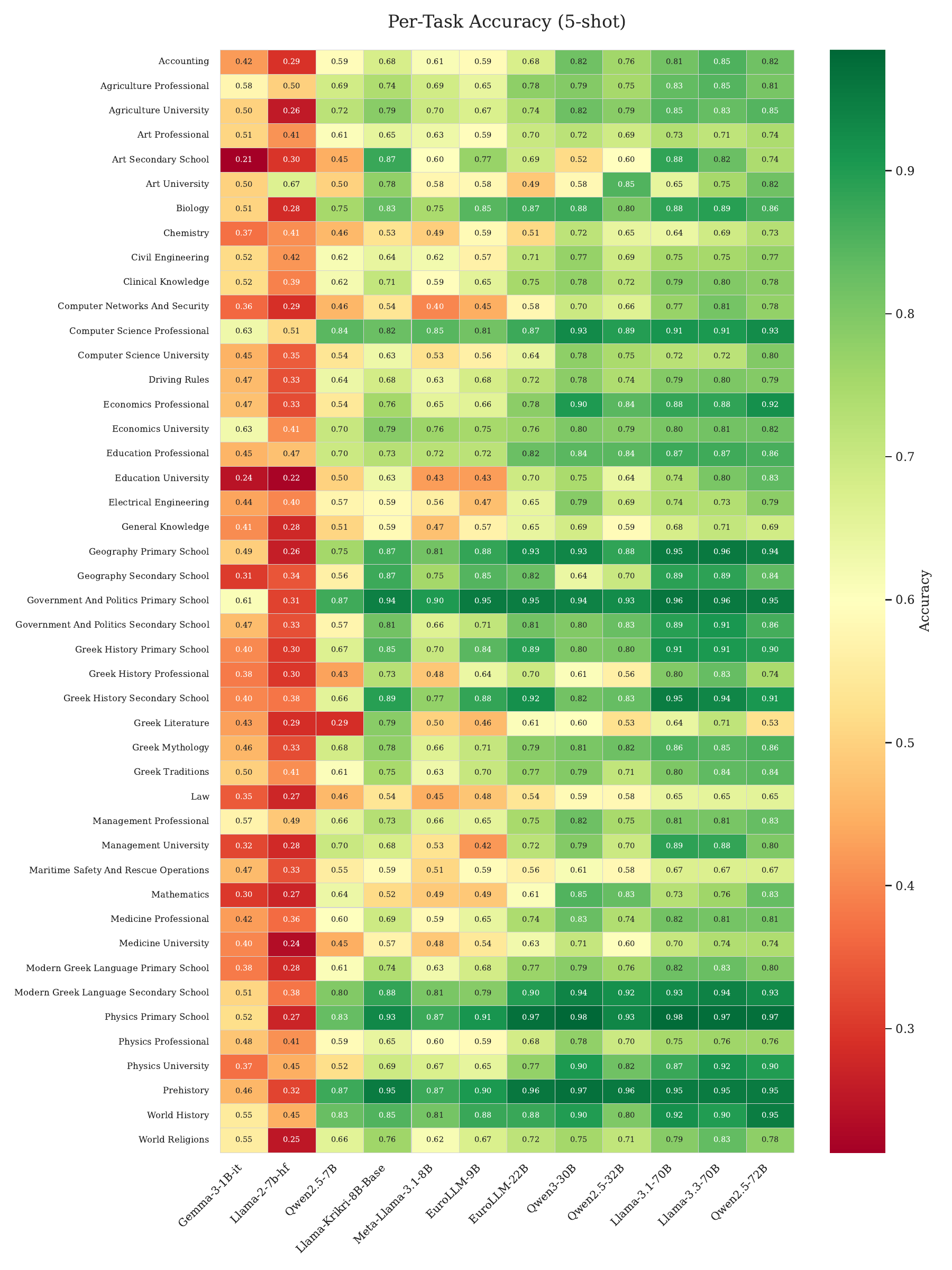}
    \caption{Subject-level accuracy heatmap under five-shot prompting, showing performance variation across GreekMMLU subjects and models.}
    \label{fig:heatmap_acc_5shot}
\end{figure*}

\begin{figure*}[h]
    \centering
    \includegraphics[width=\textwidth]{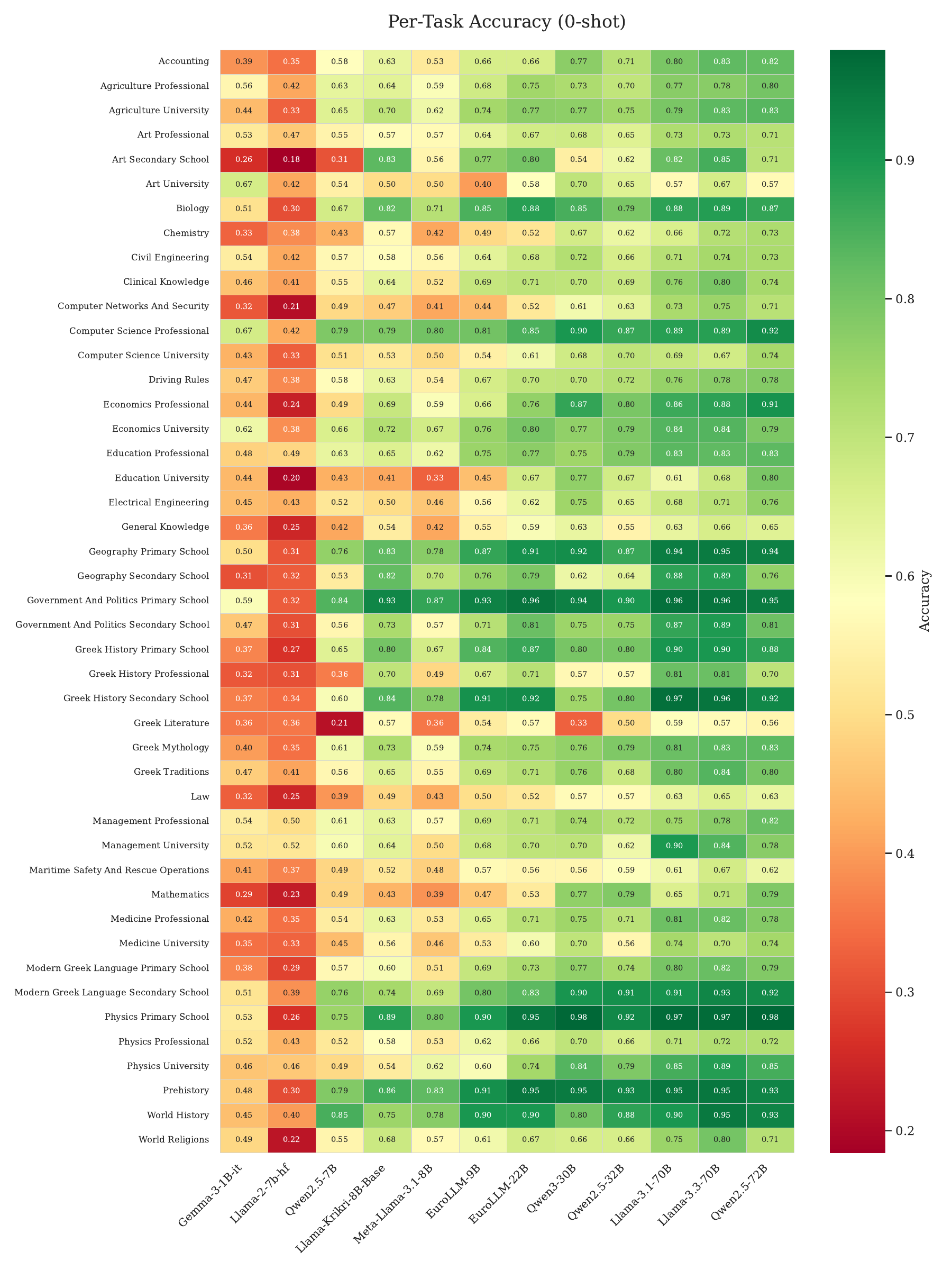}
    \caption{Subject-level accuracy heatmap under zero-shot prompting, showing performance variation across GreekMMLU subjects and models.}
    \label{fig:heatmap_acc_0shot}
\end{figure*}

As illustrated in Figures~\ref{fig:heatmap_acc_5shot} and~\ref{fig:heatmap_acc_0shot}, a subject-wise breakdown reveals heterogeneous performance patterns across the benchmark. Accuracy levels differ notably between disciplinary groups, with non-technical domains generally exhibiting narrower variance across models, while technical subjects display wider performance dispersion. Topics involving culturally grounded knowledge, including Greek historical and mythological content, introduce additional variability, particularly among general multilingual models. In contrast, models incorporating regionally focused training data tend to show more stable behavior across these subjects. 

\section{Impact of Question Length on Model Confidence} \label{sec:length_vs_confidence}
\begin{figure*}[h]
    \centering 
    \includegraphics[width=\textwidth]{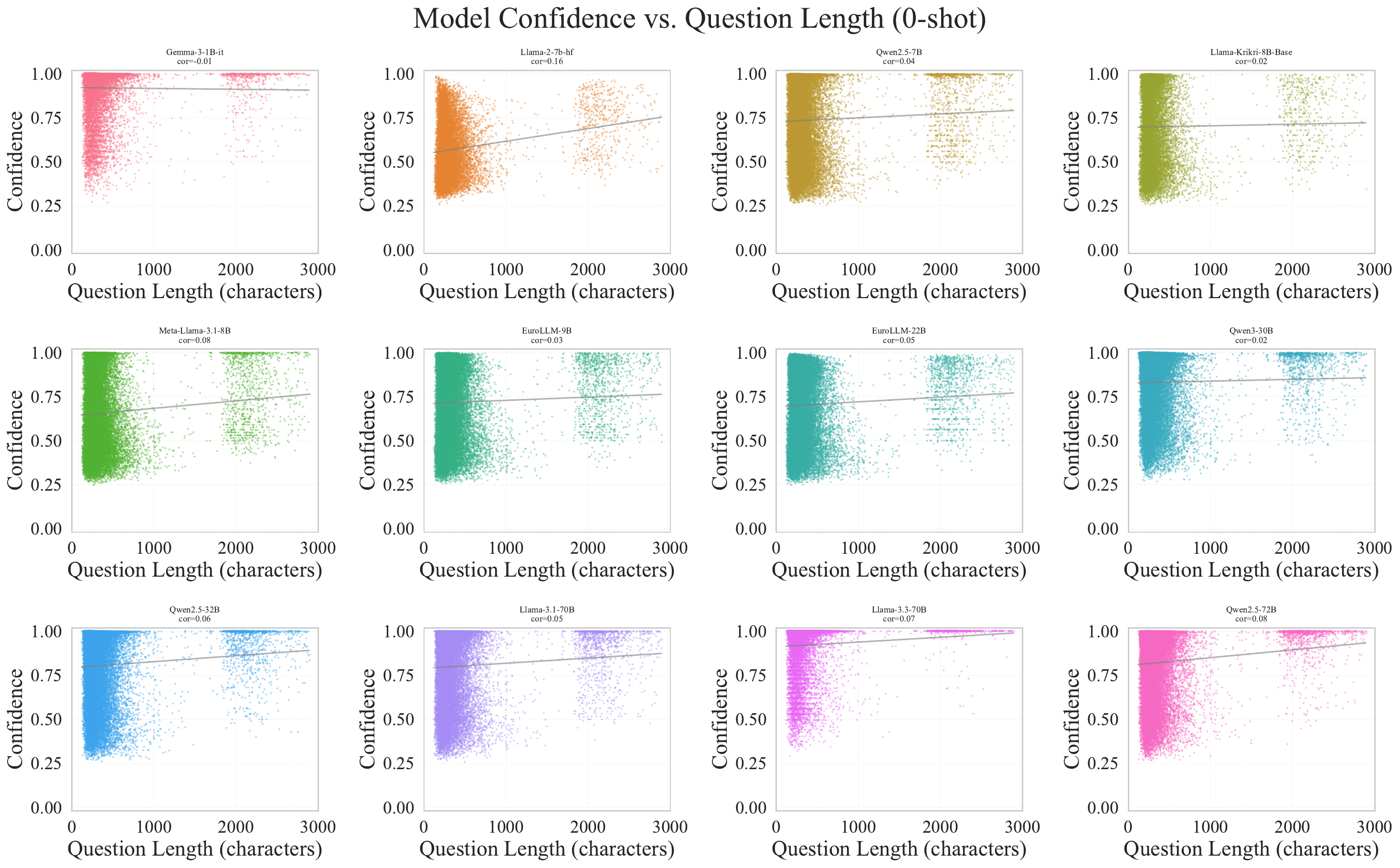} 
    \caption{Correlation between Question Length and Confidence (Zero-Shot).}  
    \label{fig:len_conf_0shot} 
\end{figure*}

\begin{figure*}[h]
    \centering 
    \includegraphics[width=\textwidth]{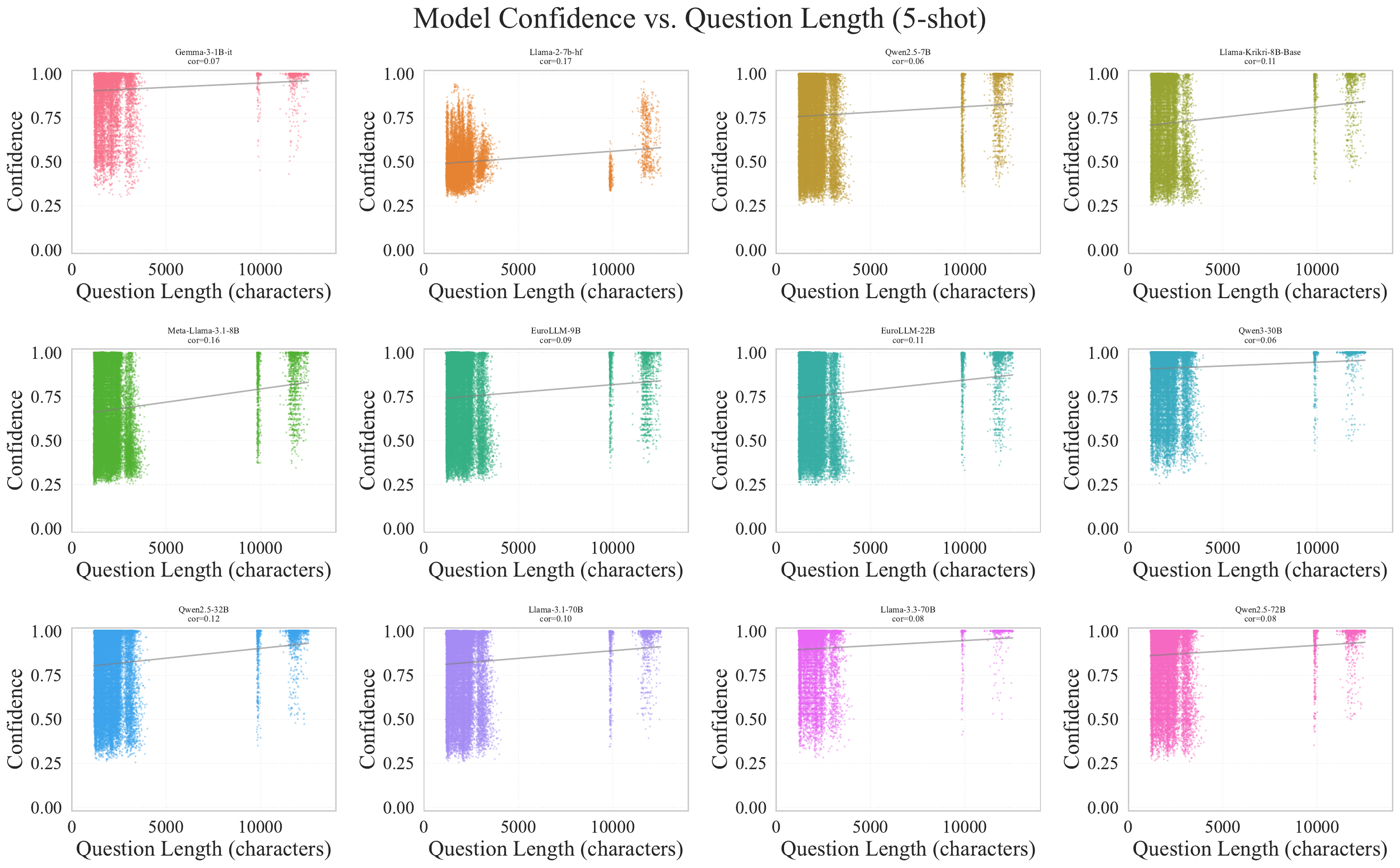} 
    \caption{{Correlation between Question Length and Confidence (Five-Shot).}}
    \label{fig:len_conf_5shot} 
\end{figure*}

To investigate whether model confidence is a byproduct of input verbosity rather than semantic certainty, we analyze the correlation between question length (measured in characters) and average confidence scores. As illustrated in Figures \ref{fig:len_conf_0shot} and \ref{fig:len_conf_5shot}, we observe no meaningful correlation between question length and model confidence for either the specialized Llama-Krikri-8B or the generic baselines (e.g., Llama-3.1-70B). Pearson correlation coefficients remain consistently close to zero across prompting strategies, indicating that the models’ uncertainty estimates are robust to variations in input length and are not driven by superficial properties of the prompt, with Llama-based models showing a marginally higher sensitivity to question length.

\end{document}